\pdfoutput=1

\documentclass[pdftex,table,xcdraw,11pt]{article}
\usepackage{microtype}
\usepackage{xspace}
\newcommand{\dataset}{\textbf{\textsc{SCRIPTS}}\xspace}
\usepackage{acl}
\usepackage{graphicx}
\usepackage{hyperref}
\usepackage{times}
\usepackage{latexsym}
\usepackage{adjustbox}
\usepackage{kotex} 
\usepackage{amsmath}
\usepackage[most]{tcolorbox}
\usepackage[T1]{fontenc}
\usepackage{tabularx}
\usepackage[utf8]{inputenc}
\usepackage[table]{xcolor}
\usepackage{microtype}
\usepackage{geometry}
\usepackage{array}
\usepackage{multirow}
\usepackage{booktabs}
\usepackage{pifont} % For special symbols
\usepackage{amssymb}
\usepackage{caption}
\usepackage{subcaption}
\usepackage{pifont}
\usepackage{amssymb}
\usepackage{stfloats}
\usepackage{makecell}
\usepackage{colortbl}
\usepackage{enumitem}

\usepackage{pgf} % for \pgfmathparse

\usepackage{fontawesome}
\usepackage[T1]{fontenc}

\usepackage{tikz}
\usetikzlibrary{arrows.meta, positioning}

\tikzset{
  box/.style = {rectangle, draw, rounded corners,
                align=left, inner sep=6pt, minimum height=8mm,
                text width=6.2cm},
  head/.style = {font=\bfseries\small, align=center},
  arrow/.style = {->, thick, >=Latex}
}

\makeatletter
\let\savedttdefault\ttdefault
\usepackage[scaled=.92]{zi4} % Inconsolata, 살짝 축소
\renewcommand{\ttdefault}{\savedttdefault}
\makeatother

\newcommand{\mytextttfamily}{\fontfamily{zi4}\selectfont}
\DeclareTextFontCommand{\texttt}{\mytextttfamily}

\newif\ifcommentsoff
\definecolor{HLikely}{HTML}{8ECAE6}
\newcommand{\hlhigh}[1]{%
  {\setlength{\fboxsep}{1pt}\colorbox{HLikely!30}{#1}}%
}
\definecolor{LLikely}{HTML}{FFB703}
\newcommand{\llhigh}[1]{%
  {\setlength{\fboxsep}{1pt}\colorbox{LLikely!30}{#1}}%
}
\definecolor{Unlikely}{HTML}{FB7185}

\newcommand{\ulhigh}[1]{%
  {\setlength{\fboxsep}{1pt}\colorbox{Unlikely!30}{#1}}%
}

\definecolor{ErrGray}{HTML}{B0B0B0}
\title{\textit{Are they lovers or friends?}\\ Evaluating LLMs' Social Reasoning in English and Korean Dialogues}

\author{%
Eunsu Kim$^{\spadesuit}$\,
Junyeong Park$^{\spadesuit}$\,
  Juhyun Oh$^{\spadesuit}$, 
  Kiwoong Park$^{\spadesuit}$, \\
  \textbf{Seyoung Song}$^{\spadesuit}$, 
    \textbf{A. Seza Doğruöz}$^{\heartsuit}$, 
  \textbf{Alice Oh}$^{\spadesuit}$,
  \textbf{Najoung Kim}$^{\diamondsuit}$ \\
  $^\spadesuit$KAIST, $^\heartsuit$LT3, IDLab, Universiteit Gent, $^\diamondsuit$Boston University
\\\texttt{\{kes0317, junyeong.park, 411juhyun, marspak, seyoung.song\}@kaist.ac.kr}\\ \texttt{as.dogruoz@ugent.be, alice.oh@kaist.edu, najoung@bu.edu} \\ 
}

\newcommand{\task}{social relationship reasoning\xspace}
\newcommand{\taskfull}{Social Relationship Reasoning\xspace}

\begin{document}
\maketitle

\begin{abstract}
As LLMs are increasingly deployed in real-world interactions, their social reasoning in interpersonal communication becomes critical. To explore their capabilities, we introduce \dataset, a 1.1k-dialogue dataset in English and Korean, sourced from movie scripts and propose a social reasoning task based on \dataset that evaluates the capacity of LLMs to infer the social relationships (e.g., friends, lovers) between speakers in each dialogue. 
Evaluating nine models on our task, current LLMs achieve around 75--80\% on the English dataset and 58--69\% in Korean, and models predict an \textsc{Unlikely} relationship in 10--25\% of responses in both languages.
Furthermore, we find that thinking models and chain-of-thought prompting provide minimal benefits for social reasoning and occasionally amplify social biases.
In sum, there are significant limitations in current LLMs' social reasoning capabilities especially for Korean, highlighting the need for efforts to develop socially-aware LLMs across languages.\footnote{Dataset: \href{https://huggingface.co/datasets/EunsuKim/SCRIPTS}{\texttt{huggingface.co/datasets/EunsuKim/SCRIPTS}}\\Code: \href{https://github.com/rladmstn1714/SCRIPTS}{\texttt{github.com/rladmstn1714/SCRIPTS}}}
% \footnote{An anonymized repository containing the dataset: \url{}}

% \noindent\centering
% \begin{adjustbox}{width=\linewidth} % ← 조절 가능: 0.85~1.0 권장
% \begin{tabular}{@{}c@{\hspace{6pt}}l@{}}
%   \raisebox{-0.25\height}{\includegraphics[height=11pt]{figures_new/huggingface_logo.png}} &
%   \href{https://huggingface.co/datasets/EunsuKim/SCRIPTS}{\texttt{huggingface.co/datasets/EunsuKim/SCRIPTS}} \\
%   \raisebox{-0.25\height}{\includegraphics[height=11pt]{figures_new/github_logo.png}} &
%   \href{https://github.com/rladmstn1714/SCRIPTS}{\texttt{github.com/rladmstn1714/SCRIPTS}} \\
% \end{tabular}
% \end{adjustbox}

\end{abstract}
\begin{figure*}[ht]
  \centering
  \includegraphics[width=\textwidth]{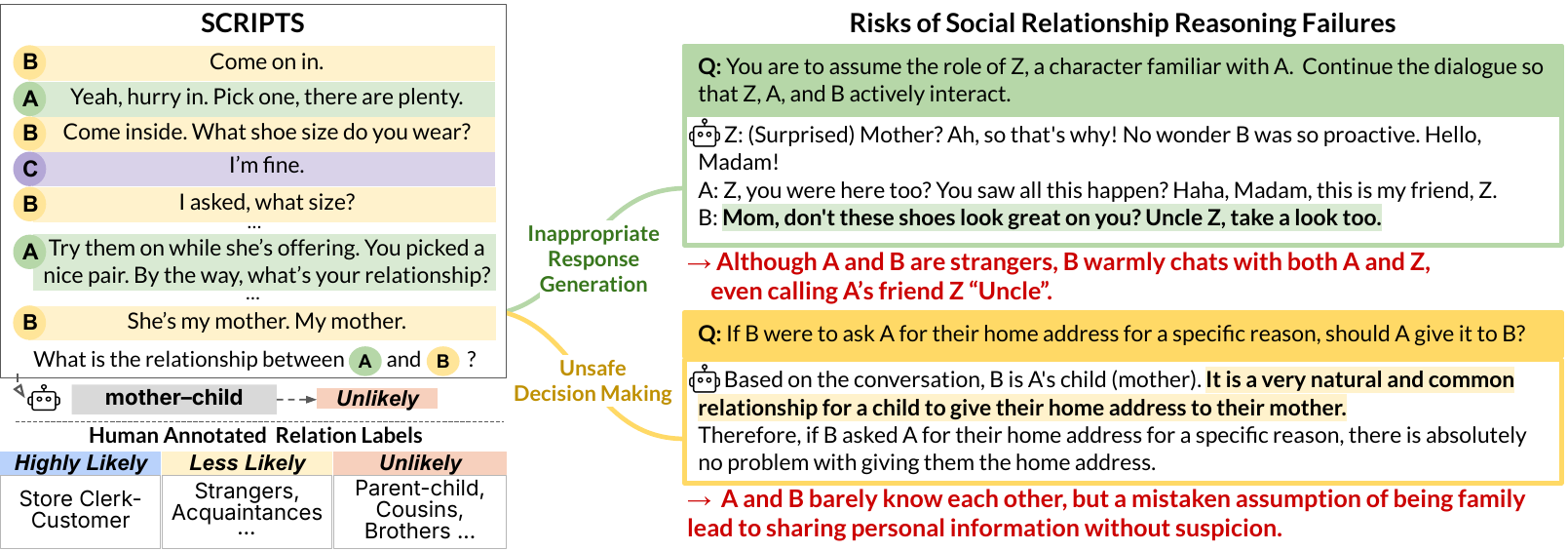}
\caption{
\textbf{Illustration of \dataset and potential risks from failures in \taskfull.} 
\dataset provides three human-annotated relationship labels: \hlhigh{\textsc{Highly~Likely}}, \llhigh{\textsc{Less~Likely}}, and \ulhigh{\textsc{Unlikely}}. 
In this example, Gemini-2.5-Flash incorrectly predicts a store clerk-customer relationship to be a mother-child relationship. 
Such misleading relational reasoning can lead to inappropriate responses and unsafe decisions, such as privacy leakage. 
Examples from Korean dialogues are translated into English for ease of reading.}

\label{fig:teaser}
\end{figure*}

\section{Introduction}
As LLM-based agents become more prevalent, we expect frequent interactions among multiple LLM agents and users~\cite{cai2024advancing,liu2024peergpt}. This trend is already reflected in practice (e.g., Group chats in ChatGPT~\cite{openai2025groupchats}), and for more natural and smooth communication, LLMs are expected to recognize the relationship between the participants \cite{sehl2022social}.
We refer to the ability to recognize and identify the relevant social relationships (e.g., lovers, friends, family members) as \textit{\task}. When LLMs fail in this type of reasoning, they risk producing responses that violate social norms or cause safety issues, as illustrated in Figure~\ref{fig:teaser}.
%.\footnote{We illustrate this potential issue in }

Although earlier studies have made some progress in evaluating LLMs’ ability to infer social relationships, they often use simplified settings that may not fully capture real-world complexity.
For instance, some work frames the task as multiple-choice classification~\cite{Jia_Huang_Q_Zhu_2021,li2023diplomat}, considers a limited taxonomy of relationship types~\cite{Jia_Huang_Q_Zhu_2021,tigunova-etal-2021-pride}, or focuses on relatively simple dyadic conversations~\cite{jurgens-etal-2023-spouse}.
Moreover, social relationship inference is often inherently uncertain and context-dependent, so a single “correct” label may be difficult to justify in many cases~\cite{hilton1995social}. For example, the remark “You never listen to me” could express a serious complaint between romantic partners or playful banter between friends depending on the broader conversational context and the language(s) used.

% Such interpretations may further diverge across languages and cultures, underscoring the need for more robust evaluation frameworks.

To address this shortcoming of previous studies, we introduce \dataset, a novel benchmark for evaluating LLMs' \task abilities, featuring an answer schema that incorporates inherent uncertainty (Figure~\ref{fig:teaser}). It contains 1.1k dialogues (580 English and 567 Korean), derived from U.S. and Korean movie scripts which are closer to realistic and culturally grounded conversations. We adopt soft labeling, where each relationship type is annotated with likelihood-based categories: \textsc{Highly~Likely}, \textsc{Less~Likely}, and \textsc{Unlikely}. By distinguishing relationship labels by likelihood, our dataset supports fine-grained evaluation that differentiates nonsensical responses from plausible but less prominent ones.

We evaluate nine widely-used models and find that current LLMs achieve moderate accuracy but still make frequent socially implausible inferences in both languages. Even the best-performing model (GPT-4o) still responds with relationships that humans annotated as \textsc{Unlikely} in 10.9\% of cases and produces \textsc{Less~Likely} responses in 10.0\%. Lower-performing models, such as Llama-3.1-8B-Instruct, make \textsc{Highly~Likely} predictions in only 41.3\% of the English dataset. Surprisingly, chain-of-thought (CoT) prompting and thinking models, which are effective for logical reasoning, provide minimal benefits for social reasoning and occasionally amplify social biases. 

In addition to quantitative analyses, we qualitatively analyze LLM failures by examining instances where models assign \textsc{Unlikely} relationship labels and identify four key failure modes in both English and Korean. Furthermore, we investigate whether providing information about the socio-demographic backgrounds (e.g., age, gender, relationship), formality, hierarchy and intimacy between human conversational partners enhances \task and find that these factors reduce \textsc{Unlikely} outputs and mitigate nonsensical responses.

In summary, our contributions are as follows:
\begin{itemize}[topsep=1pt, itemsep=0pt]
    % \setlength\itemsep{0em}
    % \item We introduce \dataset, a benchmark for evaluating LLMs' \task in English and Korean.
    \item We introduce \dataset, a benchmark for evaluating LLMs' \task, comprising 1.1k dialogues in two languages (English and Korean) with uncertainty-aware relationship labels.
    \item Evaluating nine LLMs, we find limited social reasoning ability in both languages and models often predict \textsc{Unlikely} relationships and vary substantially in identifying \textsc{Highly~Likely} relationships.
    \item Neither CoT prompting nor thinking models help much across both languages, while adding relational information shifts predictions toward more likely inferences.
\end{itemize}

% \input{tables/dialogue-type-compare}

%\section{Background \& Related Works}
\section{Related Work}
\paragraph{Evaluating \taskfull in LLMs}

% Recent evaluations of \task in LLMs largely focus on Theory of Mind and inferring mental states~\cite{jung-etal-2024-perceptions, kim2025hypothesis}.
% \revise{While important, this line of work does not evaluate the capacity to integrate social contexts into the reasoning process~\cite{hilton1995social}.}
% Our work addresses a complementary but underexplored component of social intelligence: inferring interpersonal relationships that shape interaction, accounting for context, and participants. 
Existing research on computational models for \task often adopts simplified task setups that do not fully capture the complexity of human social relations. 
% (Table~\ref{tab:dialogue-type-compare}). 
Many studies frame the problem as a classification or multiple-choice task, which makes it difficult to capture nuanced reasoning process~\cite{li2023diplomat, Jia_Huang_Q_Zhu_2021, chen-etal-2020-mpdd}. Some datasets are built from single utterances rather than multi-turn conversations, limiting contextual variation~\cite{jurgens-etal-2023-spouse}. Other studies using conversational data have methodological limitations. For instance, PRIDE~\cite{tigunova-etal-2021-pride} gathered annotations from movie summaries instead of dialogues, and \citet{rashid-blanco-2018-characterizing} 
used global relationship labels assigned to character pairs, although that label may not always be evident from a single dialogue between them.

In contrast, our benchmark uses multi-turn dialogues annotated with multiple human-inferred labels, capturing relationships as they are expressed in dialogue. Also, we evaluate our task with open-ended generation rather than fixed-choice classification, allowing a wider range of possible relationships and better reflecting the social complexity.

\paragraph{Cultural Dependency in \taskfull}
Although most studies focus on English, \task depends on linguistic and cultural context. For example, Korean relies more heavily on \textit{terms of address} and \textit{honorifics} to encode relational information in dialogue~\cite{chung2010korean,hwang1991terms}. 

% Terms of address (ToA) are words or phrases that refer directly to another person, carrying important discourse implications~\cite{hwang1991terms}. English ToA rely mainly on personal names. In Korean, these terms include kinship terms, titles, and pronouns. For instance, instead of using an adult's name, Korean speakers may prefer ``someone else's dad,'' reflecting cultural norms that avoid direct name use in certain contexts.
% % , with limited titles (Mr., Mrs.), professional markers (Professor), and kinship terms (Mom, Dad).
% Honorifics encode social relationships into linguistic forms~\cite{ide1989formal}. While many languages employ politeness strategies, Korean has an extensive grammaticalization of honorifics~\cite{harada1976honorifics,brown2014phonetics}, whereby roles, status and formality are conveyed through verbal morphology~\cite{fukada2004universal,brown2015honorifics,pizziconi2011honorifics}. English, by contrast does not have this system. 
% % Instead, politeness is expressed through speech acts or pragmatic choices. 
% Given these cultural and linguistic differences, \task should be evaluated cross-linguistically and cross-culturally.
Terms of address (ToA) are expressions used to directly refer to another person and carry discourse and social meaning~\cite{hwang1991terms}. In English, ToA are mostly people’s personal names, whereas in Korean, they commonly include kinship terms, titles, and pronouns. For instance, instead of addressing an adult by using their first name, speakers may say ``{their child's first name}'s dad,'' (e.g., Minsu's Dad), reflecting norms that discourage direct name use in certain contexts.
While many languages encode politeness, Korean has a highly grammaticalized honorific system~\cite{harada1976honorifics,brown2014phonetics} that conveys roles, status, and formality through verbal morphology~\cite{fukada2004universal,brown2015honorifics,pizziconi2011honorifics}, unlike English which lacks an equivalent system.
These cultural and linguistic differences motivate evaluating \task cross-linguistically and cross-culturally.
% In that sense, our dataset is unique for covering two languages and cultures.
% For example, the verb “편찮으시다” (to be ill, polite form) contrasts with the plain form “아프다,” while the morpheme “-요” further marks politeness:
% \begin{enumerate}
% \small
% \setcounter{enumi}{0}
%     \item 아주머니가 [\textbf{편찮으시다}]는 소식 방금 들었어[\textbf{요}]. \\
%     I just heard that she is sick.
% \end{enumerate}

% \paragraph{Evaluating Tasks with Multiple Answers of Varying Plausibility}
% %add UQ
% Unlike domains with verifiable answers (e.g. mathematics), \task is inherently subjective~\cite{hilton1995social}, operating on a spectrum of plausible interpretations across contexts and participants. A robust evaluation should credit likely answers and penalize socially inappropriate ones. Error severity-aware evaluations are common in ambiguous NLP tasks like machine translation~\cite{freitag-etal-2021-experts} and dialogue generation~\cite{zhang-etal-2022-fined}, yet remain largely absent in \task. Prior work acknowledges ambiguity by allowing multiple labels~\cite{jurgens-etal-2023-spouse,tigunova-etal-2021-pride,yu-etal-2020-dialogue} but does not model their probabilistic nature. We extend these paradigms by separating plausible secondary interpretations from outright failures, enabling more granular evaluation.

\section{\dataset: Evaluating LLMs' Interpersonal Social Reasoning} %dataset/task 이름 짓기 

We introduce \dataset, a benchmark for \task  in multi-turn dialogues in English and Korean. 
In this section, we outline the motivation (\S~\ref{sec:dataset_motiv}), design (\S~\ref{sec:dataset_design}), and construction (\S~\ref{sec:dataset_const}) of \dataset.

\subsection{The Importance of \taskfull in LLMs}
\label{sec:dataset_motiv}
To participate in naturalistic social conversations, LLMs must produce utterances that are appropriate for the underlying relationship and context. The right side of Figure~\ref{fig:teaser} illustrates how misinterpreting relationships can lead to undesirable outcomes (e.g., social harm). In the example, an LLM (Gemini-2.5-Flash) mistakes a store clerk-customer interaction for a mother-child relationship, resulting in an inappropriate next response and potentially encouraging oversharing of personal information. 

\subsection{Dataset Design}
\label{sec:dataset_design}
To capture the complexity and diversity of real-world social dynamics, we leverage movie scripts that contain natural human interactions spanning a wide range of relationships (Table~\ref{tab:initial-relationships}). \dataset makes two key contributions: (1) it aims to capture relationships as they are contextualized within the dialogue, rather than relying on static role labels, and (2) it adopts a three-tier probabilistic labeling scheme—\hlhigh{\textsc{Highly~Likely}}, \llhigh{\textsc{Less~Likely}}, and \ulhigh{\textsc{Unlikely}}—for more fine-grained evaluation of \task.

\paragraph{Dialogue-Level Labeling} 
A key design choice of \dataset is to label relationships as they appear in specific conversational contexts. Prior benchmarks often assign fixed character roles from movie metadata (e.g., mother–son)~\cite{Jia_Huang_Q_Zhu_2021}. However, social relationships are dynamic, as speakers can shift roles across contexts and may hold multiple relationships simultaneously.

We highlight the value of dialogue-level labeling by comparing our annotations with static, movie-level labels. We find that 19\% of movie-level labels are judged irrelevant for a given dialogue (suggesting that a global label can be misleading) and even when a movie-level label is applicable, our annotations identify more than three \textsc{Highly~Likely} relationships per dialogue on average. These findings show that a single conversation often reflects multiple social facets, motivating our context-aware, dialogue-level labeling approach.

\paragraph{Probabilistic Labeling}
As social situations are inherently ambiguous, one dialogue may suggest multiple relationships with varying plausibility (Figure~\ref{fig:teaser}). Our three-tier labeling scheme (\textsc{Highly~Likely}, \textsc{Less~Likely}, and \textsc{Unlikely}) is intended to capture this. Specifically, this design has two key properties: (1) \textbf{finer granularity}, allowing metrics to reward models for identifying the most salient relation (\textsc{Highly~Likely}) rather than just plausible ones (\textsc{Less~Likely}); and (2) a \textbf{nonsense penalty}, which penalizes contextually inappropriate predictions (\textsc{Unlikely})—a critical failure in \task.

\subsection{Dataset Construction}
\label{sec:dataset_const}
\begin{table}

\centering
\resizebox{\columnwidth}{!}{%
  \begin{tabular}{l|rr|r}
    \toprule
    Type& \textbf{English} & \textbf{Korean} & \textbf{Total} \\
    \midrule
    Movies                        & 28    & 32    & 60 \\
    Dialogues                           & 580   & 567   & 1,147 \\
    3-Person Dialogues             & 223 & 256 & 479 \\
    Unique Highly-Likely Relationships  & 230   & 617   & -- \\
    Turns (avg \# per dialog)                          & 10.21 & 9.89  & 10.05 \\
    Highly-Likely Relationships (avg \# per dialog) & 3.62  & 3.72  & 3.67 \\
    Unlikely Relationships (avg \# per dialog)   & 18.50 & 23.13 & 20.79 \\
    \bottomrule
  \end{tabular}
}
\caption{\textbf{Statistics of \dataset}. SCRIPTS contains 1.1K English and Korean dialogues from 60 movies, annotated with 230+ unique relationship types.}
\label{tab:basic-stats}
\end{table}

% \begin{table*}[t]
% \resizebox{\textwidth}{!}{%
% \begin{tabular}{|l|l|}
% \hline
% Category   & Specific Relationships                          \\ \hline
% Family         & Parent-Children, Brothers/Sisters, Grandparent-Grandchildren, Cousins, Uncle/Aunt-Niece    \\ \hline
% Social     & Friends, Acquaintances, Neighbors, Strangers    \\ \hline
% Romance        & Romantic Interest, Dating, Married, Engaged, Friends wth benifits, Affair, Ex-relationship \\ \hline
% Organizational & Coworkers, Professonal colleagues, Supervisor-Subordinate relationship                     \\ \hline
% Role-based     & Mentor-Mentee, Teacher-Student, Lawyer-Client, Doctor-Patient, Landlord-Tenant             \\ \hline
% Antagonist & Competitive relationship, Rivalry, Arch-enemies \\ \hline
% \end{tabular}%
% }
% \caption{Initial Relationships}
% \label{tab:initial-relationships}
% \end{table*}

% \usepackage{booktabs}
% \usepackage{graphicx}
\begin{table}[ht]
\centering
\resizebox{0.9\columnwidth}{!}{%
\begin{tabular}{@{}ll@{}}
\toprule
\textbf{Category} & \textbf{Specific Relationships}                                                           \\ \midrule
Family         & \begin{tabular}[c]{@{}l@{}}Parent-Children, Brothers/Sisters,\\ Grandparent-Grandchildren, Cousins,\\ Uncle/Aunt-Niece\end{tabular}    \\ \midrule
Social                   & \begin{tabular}[c]{@{}l@{}}Friends, Acquaintances, Neighbors,\\ Strangers\end{tabular}    \\ \midrule
Romance        & \begin{tabular}[c]{@{}l@{}}Romantic Interest, Dating, Married,\\ Engaged, Friends wth benifits, Affair,\\ Ex-relationship\end{tabular} \\ \midrule
Organizational & \begin{tabular}[c]{@{}l@{}}Coworkers, Professonal colleagues,\\ Supervisor-Subordinate relationship\end{tabular}                       \\ \midrule
Role-based     & \begin{tabular}[c]{@{}l@{}}Mentor-Mentee, Teacher-Student,\\ Lawyer-Client, Doctor-Patient,\\ Landlord-Tenant\end{tabular}             \\ \midrule
Antagonist               & \begin{tabular}[c]{@{}l@{}}Competitive relationship, Rivalry,\\ Arch-enemies\end{tabular} \\ \bottomrule
\end{tabular}%
}
\caption{\textbf{Initial relationships used for \ulhigh{\textsc{Unlikely}} annotation (27 items)}.}
\label{tab:initial-relationships}
\end{table}
We collect 60 movie scripts: 28 English scripts crawled from IMSDb and 32 Korean scripts obtained via an onsite visit to the Korean Film Archive and crawling an open-access Korean script community.\footnote{
\href{https://imsdb.com}{imsdb.com};
\href{http://www.kmdb.or.kr}{kmdb.or.kr};
\href{https://www.filmmakers.co.kr}{filmmakers.co.kr}.
} The full movie list and metadata are provided in Appendix Table~\ref{tab:movies}.

We filter scenes to those with at least three turns ($\geq$ 3) and two or three participants. Among the remaining scenes, we prioritize those with diverse speaker combinations (i.e., avoiding repeated exchanges among the same few participants). From 23k initial scenes, this yields 1,322 high-quality dialogues (698 English; 624 Korean) for human annotation. Additional collection and filtering details are in Appendix~\ref{appendix:dataset}.
While prior work primarily studies dyadic interactions, \dataset\ includes three-speaker dialogues (41.8\%) to capture more complex social settings. For these scenes, the task remains dyadic relationship inference between two interlocutors and we randomly select the speaker pair.

\subsubsection{Collecting Human Annotations}
\label{sec:method_human_annotation}
Each dialogue is annotated by three annotators who are native or near-native speakers with extensive cultural familiarity (e.g., 10+ years of residence in U.S. and South Korea, respectively). We recruit 17 annotators for English and 14 annotators for Korean (see Appendix~\ref{appendix:Human Annotation} for details).

Our annotation procedure is designed to preserve uncertainty in social relationship inference while filtering obvious annotation noise. To this end, we treat \textsc{Unlikely} relationships conservatively through majority agreement, while allowing multiple \textsc{Highly Likely} relationships to remain in the final label set. The full annotation interface and instructions are provided in Appendix~\ref{appendix:integrating_social_info}.

\paragraph{Phase 1: Labeling \ulhigh{\textsc{Unlikely}} Relationships}
Annotators are shown a predefined set of 27 relationship types compiled from previous work (Table~\ref{tab:initial-relationships}). Annotators select all relationships that are clearly contradicted by the dialogue. Because these labels are used as high-confidence negatives, a relationship is assigned to the \textsc{Unlikely} set only if at least two of the three annotators select it.

\paragraph{Phase 2: Open-ended Labeling for \hlhigh{\textsc{Highly~Likely}} Relationships}
Annotators provide open-ended text labels for the relationship(s) most strongly supported by the dialogue. We use open-ended annotation here because the socially most plausible relationship is often context-dependent and may not be well captured by predefined categories. Annotators can provide up to five relationship labels for each dialogue. We normalize annotators' responses to a standardized label space (e.g., merging variants such as ``mother-child'' and ``mom-child'' and mapping open-ended labels to corresponding predefined categories when applicable). After normalization, we take the union of all annotators' \textsc{Highly~Likely} labels as the final \textsc{Highly~Likely} set for the dialogue. 

\paragraph{Phase 3: Deriving \llhigh{\textsc{Less~Likely}} Relationships}
We define \textsc{Less~Likely} labels as the remaining relationship types from the predefined set that are neither included in the final \textsc{Highly~Likely} set nor marked as \textsc{Unlikely}. 

\paragraph{Phase 4: Annotating Auxiliary Labels}
Annotators additionally label socio-demographic attributes (e.g., age, gender) and relational dimensions (e.g., formality, hierarchy, intimacy). We use these auxiliary labels for downstream analyses of which social cues support relationship inference~\cite{wish-perceive-1976,nguyen-etal-2016-survey}.
% Humans interpret social relationships considering sociolinguistic information and relational dimension

\paragraph{Quality Control}
To ensure annotation quality, we run training sessions and a pilot study. We exclude dialogues where the three annotators’ \textsc{Highly~Likely} labels have no overlap (i.e., are mutually exclusive), indicating low reliability. This yields 1,147 dialogues (580 English; 567 Korean), removing 13.2\% of the initial dataset. We report agreement separately by annotation type in Appendix~\ref{appendix:Human Annotation}. 
%Table~\ref{tab:basic-stats} reports the statistics of the final dataset. 

Table \ref{tab:basic-stats} shows the detailed statistics of \dataset. Also, see Appendix~\ref{appendix:dataset_diversity} for comparisons of the types and frequencies of relationships in English and Korean dialogues.

% \subsection{Diversity of \dataset}
% \label{sec:method-diversity}

% As shown in Table~\ref{tab:basic-stats}, our dataset comprises more than 230 unique relationships in English and 617 in Korean. On average, each dialogue is annotated with 3.7 \textsc{Highly~Likely} relationships and 20.8 \textsc{Unlikely} relationships. Further statistical comparisons between English and Korean datasets are provided in Appendix~\ref{appendix:dataset_diversity}.

% Figure~\ref{fig:main_probable_subd_dist_en} illustrates that our dataset can capture diverse interpersonal dynamics by labeling relational dimensions. 
% The typicality of certain relationships is often defined by their levels of intimacy, formality, and hierarchy~\cite{wish-perceive-1976}. For instance, friendship is generally characterized as intimate, non-hierarchical (equal), and informal. Yet, our dataset also includes atypical relationships. For instance, over 40\% of friend relationships in our dataset deviate from these typical dimensions.

\section{Evaluating LLMs with \dataset}
\label{sec:exp}

\begin{figure*}[ht]
  % 첫 번째 서브피겨
  \begin{subfigure}[b]{0.48\textwidth}
    \centering
    \includegraphics[width=\textwidth]{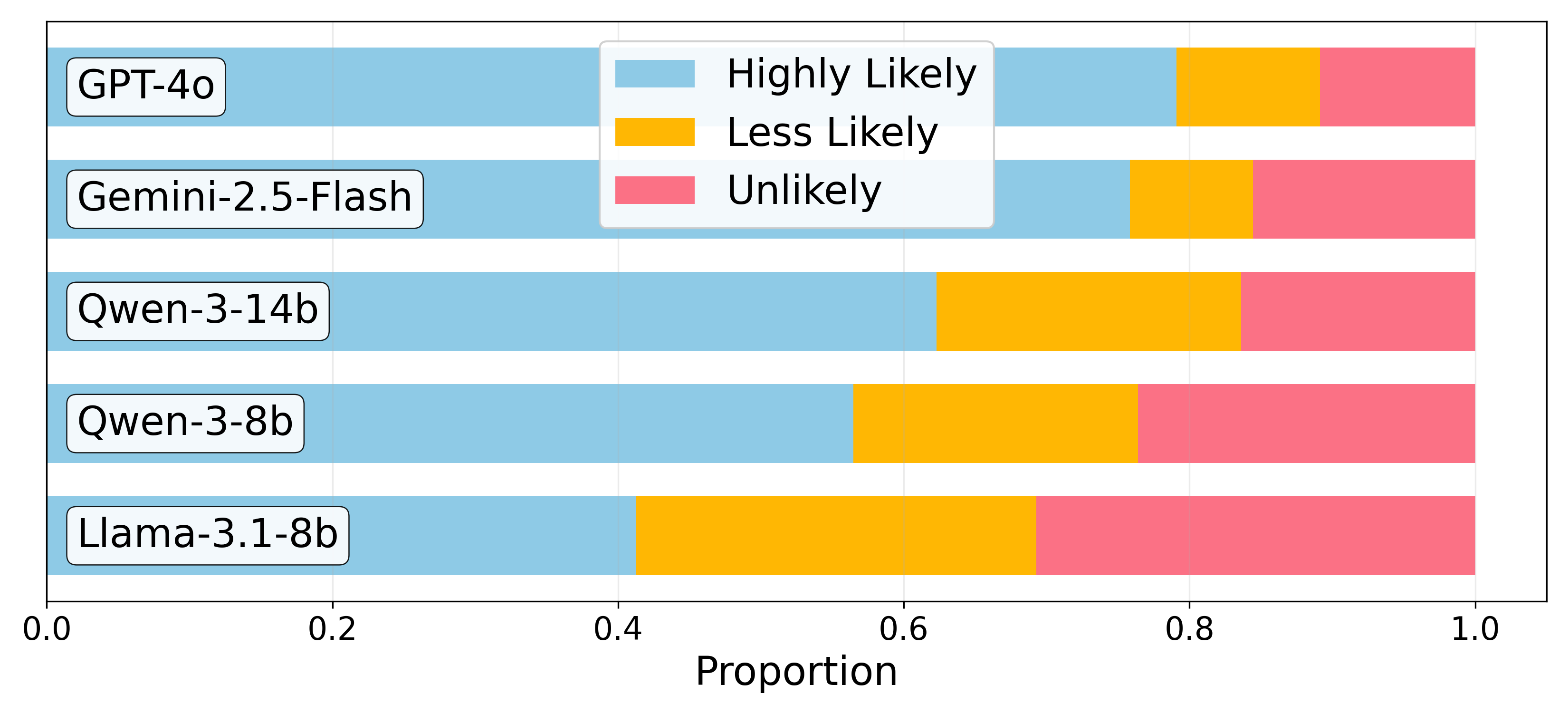}
    \caption{English}
    \label{fig:en_main}
  \end{subfigure}
    \hfill
  % 두 번째 서브피겨
  \begin{subfigure}[b]{0.48\textwidth}
    \centering
    \includegraphics[width=\textwidth]{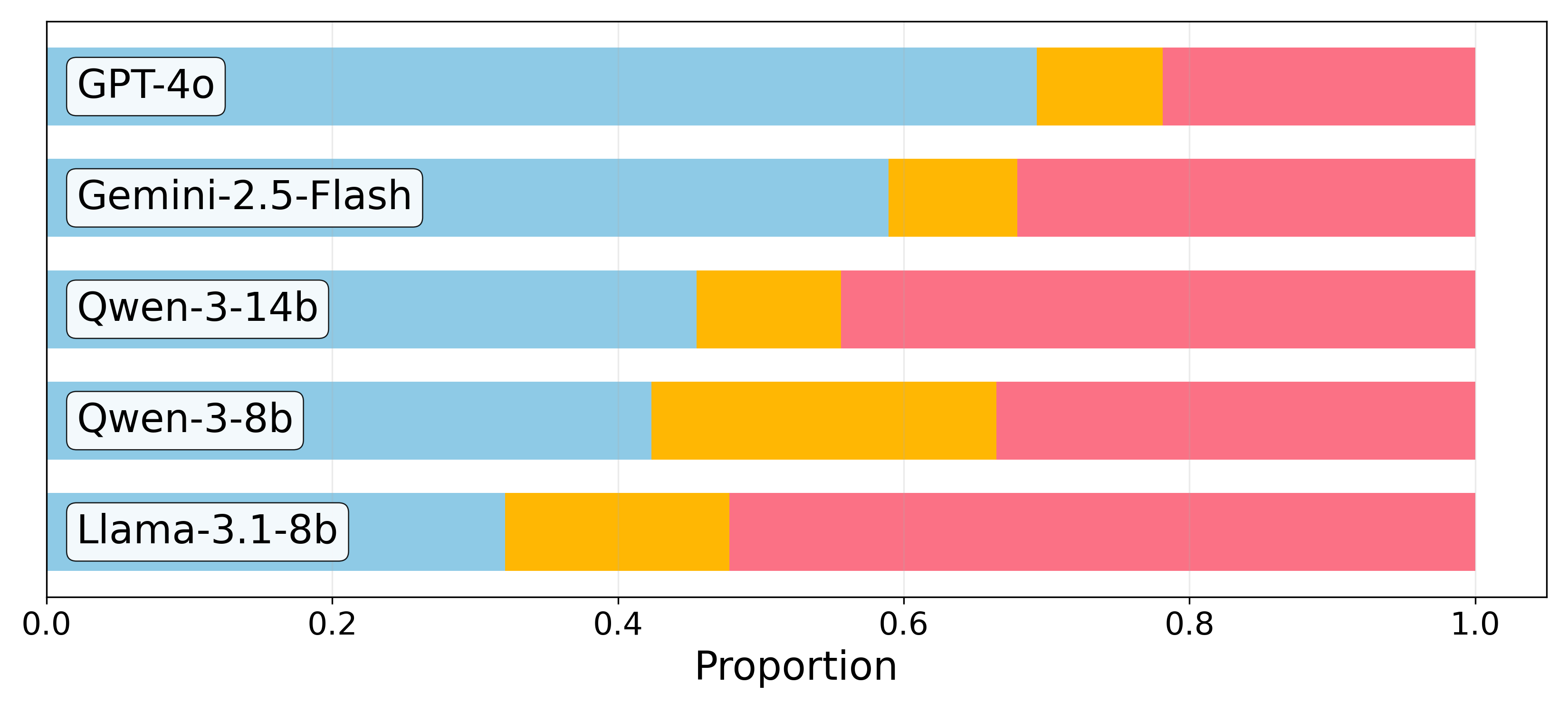}
    \caption{Korean}
    \label{fig:ko_main}
  \end{subfigure}
  
  \caption{\textbf{Comparison of model performance in English and Korean datasets.} \hlhigh{\textsc{Highly~Likely}} represents
the accuracy of the model’s majority response being a highly likely relationship, while \ulhigh{\textsc{Unlikely}} indicates the error
rate where an unlikely relationship is included in the model responses. \llhigh{\textsc{Less~Likely}} indicates the proportion of cases in which the model generates neither a \textsc{Highly~Likely} nor an \textsc{Unlikely} prediction.}
  \label{fig:main_breakdown}
  \end{figure*}
We evaluate 9 LLMs: 3 proprietary models~(GPT-4o, o3, Gemini-2.5-flash), 3 widely used open-source models~(Qwen-3-\{8B/14b\}, Llama-3.1-8B-instruct), and 3 open-source multilingual models specialized for Korean~(A.X-4.0-light-7B, Kanana-1.5-8B, and Exaone-4.0-30B).\footnote{See Appendix~\ref{appendix:exp_model_config} for model configurations.} We additionally report results for five more models—GPT-4.1, GPT-5, Qwen3-32B, Llama3.3-70B-Instruct, and Kanana2-30B-A3B-Instruct—in Appendix~\ref{appendix:exp-result}.
\begin{tcolorbox}[
    enhanced,
    breakable,
    colframe=gray!60,
    colback=gray!8,
    boxrule=0.4mm,
    arc=3.5mm,
    outer arc=3.5mm,
    title={Inference Prompt (EN)},
    fonttitle=\bfseries,
    coltitle=black,
    left=2mm,
    right=2mm,
    top=1mm,
    bottom=1mm
]
\small
Read the following conversation and guess the relationship of the participants [A] and [B].  
When guessing the relationship, refer to the following examples of relationships:
\texttt{\{example\_relations\}}

If the relationship matches one of the examples above, use it as is,  
but if the relationship does not fit any of the examples, describe the relationship yourself.  

Your answer about the relationship must be in JSON format:
\begin{verbatim}
{"relation": ""}
\end{verbatim}

\textbf{Conversation:}\\
\texttt{\{dialogue\}}

\textbf{Output (JSON):}
\end{tcolorbox}
We evaluate models in an open-ended generation setting: given a dialogue, each model generates the social relationship(s) between the target speakers in free-form text rather than selecting from a fixed label set. The prompt includes example relationship types from prior work (Table~\ref{tab:initial-relationships}) as reference candidates, while still allowing models to generate relationships outside the list.

\paragraph{Evaluation and Metrics}
Considering the probabilistic nature of the task, we run each model five times per dialogue and take the model’s majority response among them. We then compute (1) the proportion of samples whose majority response falls into the \textsc{Highly~Likely} relation set and (2) the proportion of which majority response falls into the \textsc{Unlikely} relation set. We use GPT-4o to evaluate each model's short-form answers based on ground-truth labels, following common practice in prior LLM evaluation work. In our validation experiment, GPT-4o as an evaluator yields 92.0\% human-validated accuracy (Appendix~\ref{appendix_exp_validate_judge}).

\subsection{Overall Performance}
Figure~\ref{fig:main_breakdown} shows the performance of five multilingual models (i.e., GPT-4o, Gemini-2.5-Flash, Qwen-3-\{8/14\}B, and Llama-3.1-8B-Instruct).
We find that GPT-4o achieves the best performance with \textsc{Highly~Likely} rate of 79\% and 69\% in English and Korean, respectively. The models incorrectly infer an \textsc{Unlikely} relationship in 10.8--31.9\% of their responses and this tendency is amplified in the Korean dataset with a rate increasing by an additional 7.2--16.5\%p. Table~\ref{tab:main_results} in Appendix~\ref{appendix:exp} provides the exact numerical values. We provide a case study of these behaviors with the frequent failure modes of the models in \S\ref{sec:qual-eval}.

\subsection{Does Thinking Help?}
\label{sec:exp_think}
% {Performance of Models with CoT prompting and Thinking Processes}

\begin{table*}[ht!]
\centering
\small
\begin{tabular}{lccccc}
\toprule
Model & Thinking & \multicolumn{2}{c}{En} & \multicolumn{2}{c}{Ko} \\
\cmidrule(lr){3-4} \cmidrule(lr){5-6}
 &  & \hlhigh{\textsc{Highly~Likely} (↑)} & \ulhigh{\textsc{UnLikely} (↓)}  &  \hlhigh{\textsc{Highly~Likely} (↑)} & \ulhigh{\textsc{UnLikely} (↓)}  \\
\midrule
OpenAI/GPT-4o             & $\times$ & 0.767 & 0.116 & 0.642 & 0.215 \\
OpenAI/o3                 & \checkmark & \textbf{0.807} & \textbf{0.086} & \textbf{0.742} & \textbf{0.152}\\ \midrule
Gemini-2.5-flash   & $\times$ & 0.759 & 0.154 & \textbf{0.582}&  0.318 \\
Gemini-2.5-flash   &  \checkmark  &\textbf{ 0.776} &\textbf{ 0.138} & 0.538 & \textbf{0.239} \\ \midrule
Qwen-3-14b         & $\times$ &0.623    & 0.164    & 0.455    & 0.444    \\
Qwen-3-14b         &  \checkmark  & \textbf{0.673}   & \textbf{0.107}   & \textbf{0.467}    & \textbf{0.443}    \\
\bottomrule
\end{tabular}
\caption{\textbf{Model comparison with and without Thinking mode across English (En) and Korean (Ko).} }
\label{tab:model_thinking_comparison}
\end{table*}

We analyze the performance of models when CoT prompting or internal thinking processes are incorporated. These methods have been shown to be effective for improving reasoning on math and scientific tasks~\cite{wei2022chain}.

\paragraph{Effectiveness of Chain of Thought Prompting} We apply CoT prompting on four multilingual models (one per family): GPT-4o, Gemini 2.5 Flash, Qwen-3-8B, and Llama-3.1-8B-Instruct. As shown in Table~\ref{tab:model Cot}, CoT does not consistently help: Gemini 2.5 Flash shows a 1.7\%p drop in \textsc{Highly~Likely} responses in English, and Llama-3.1-8B-Instruct shows a 3.1\%p rise in \textsc{Unlikely} responses in Korean. This contrasts with other types of reasoning tasks (e.g., math), where CoT often helps, suggesting that social reasoning requires a fundamentally different reasoning strategy.

\paragraph{Effectiveness of Thinking Process}
We evaluate three reasoning models: o3, Gemini-2.5-Flash, and Qwen-3-14B, comparing the latter two with and without an internal thinking process. As o3 does not support disabling its internal thinking process, we instead compare it with GPT-4o, a non-thinking model from the same provider. Due to budget constraints, we run each model only once per dialogue, unlike the setting used for Figure~\ref{fig:main_breakdown}, where each model is run five times per dialogue. As shown in Table~\ref{tab:model_thinking_comparison}, enabling thinking yields mixed results across languages. In English, thinking sometimes leads to slightly higher performance, whereas in Korean its effect is negligible and can even hurt performance (e.g., a 4.4\%p drop in \textsc{Highly~Likely} responses for Gemini-2.5-Flash). However, none of these differences are statistically significant (bootstrap test, $p > 0.05$). Overall, thinking does not provide a meaningful advantage for this task.
\subsection{Do Korean-specialized models perform better on Korean dialogues?}

\begin{table}[t]
\centering
\small
\resizebox{0.95\columnwidth}{!}{%
\begin{tabular}{lcc}
\toprule
Rank & En & Ko \\
\midrule
\textbf{1} & \textbf{A.X-4.0-Light (0.589)} & \textbf{A.X-4.0-Light (0.467)} \\
\textbf{2} & \textbf{ Qwen-3 (0.565)}      & \textbf{Qwen-3 (0.423)} \\
\textbf{3} & \textbf{Llama-3.1 (0.413)}  & Exaone-4.0 (0.409) \\ 
\textbf{4} & Kanana-1.5 (0.406)                     & Kanana-1.5 (0.328) \\
\textbf{5} & Exaone-4.0 (0.318)                     & \textbf{Llama-3.1 (0.321)} \\
\bottomrule
\end{tabular}
}
\caption{\textbf{Model ranking of Korean-specialized and open-source models in English and Korean}, based on the \hlhigh{\textsc{Highly~Likely}} response rate (numbers in parentheses indicate the corresponding values).}
\label{tab:main_ko_en_ranking}
\end{table}

We evaluate three Korean-specialized models: A.X-4.0-light (7B), Kanana-1.5-8B, and Exaone-4.0-32B. A.X-4.0\footnote{\href{https://github.com/SKT-AI/A.X-4.0/}{https://github.com/SKT-AI/A.X-4.0/}}
 is further trained on Korean data on top of Qwen. Kanana’s technical report also indicates stronger Korean performance and competitive English performance relative to other models across various benchmarks.~\cite{kananallmteam2025kananacomputeefficientbilinguallanguage}.

% https://tech.kakao.com/posts/716

Table~\ref{tab:main_ko_en_ranking} compares the three models with similarly sized open-source multilingual models (Qwen-3-8B, Llama-3.1-8B-Instruct). The results show that A.X-4.0-Light and Qwen-3-8B achieve the best and second-best performance in both languages, but the 3rd–5th rankings differ. In English, Llama-3.1-Instruct-8B ranks 3rd, while in Korean, Exaone-4.0-32B and Kanana-1.5-8B, take 3rd and 4th, with Llama-3.1-Instruct-8B ranking last. Full results are in Appendix~\ref{appendix:result_kmodels} (Table~\ref{tab:k_model_full}).

\section{Reasons Behind Failure of LLMs}
\label{sec:qual-eval}
Based on qualitative analyses of the reasoning processes of LLMs in CoT experiments (\S\ref{sec:exp_think}), we identify four types of failures.
\paragraph{Failure to Distinguish Terms of Address and Reference}
Models often misinterpret a term of reference (i.e., a word used to refer to someone) as a term of address (i.e., a word used to call someone directly), leading to a fundamental misunderstanding of the social context.

\begin{tcolorbox}[breakable,
    colback=gray!5!white,  
    colframe=gray!50!gray,
    fonttitle=\bfseries,  
    arc=1mm,              
    boxrule=0.5pt,       
    fontupper=\scriptsize,
    boxsep=1pt,          
    left=2pt,             
    right=2pt,            
    top=2pt,              
    bottom=2pt            
]
{\scriptsize
\textbf{Dialogue 1 (English):}
\vspace{-4pt} 
\setlength{\leftmargini}{4pt}
\begin{quote}
\textbf{[A]:} Hi Officer, can I help you? \\
\textbf{[B]:} Yes, I'm hoping you can. An elderly gentleman went missing from the nursing home down the street. Staff seems to think he came here. \\
(...)\\
\textbf{[A]:} (Pause, then) Oh....that's my Dad. He can't talk. Had a major stroke a few years back. But he's doing well. Ain't ya Pop? \\
\textbf{[B]:} OK, well, thanks for your time. Here's my number in case you hear of anything. Sorry to bother you. \\
(...)
\end{quote}
\vspace{-4pt} 
\noindent\textbf{Ground Truth:} Police officer--Civilian, Strangers \\
\textbf{Prediction:} Parent--Children / Father--son (Llama, Gemini, GPT)
}
\end{tcolorbox}

In Dialogue 1, speaker [A] says, \textit{``that's my Dad''}, using \textit{``Dad''} as a reference to identify a third person for the police officer [B]. However, the models latch onto this keyword and misinterpret it as a term of address from [A] to [B], leading to the incorrect inference of a Parent-Child relationship. This failure leads the models to ignore clear cues (e.g., [B] being called \textit{``Officer''}) that contradict this interpretation.

In Korean, this error is more pervasive because speakers often use terms of address to refer to themselves. For example, a teacher may tell a student, “The teacher (I) told you to do this,” where “the teacher” is a self-reference. However, models may misread “the teacher” as a third party.

\paragraph{Failure to Aggregate Multiple Cues}
Social relationship reasoning requires the ability to identify multiple cues within the context and integrate them to arrive at a conclusion.
However, models fail to integrate cues, especially when their combination is atypical.

\begin{tcolorbox}[breakable,
    colback=gray!5!white,   % 박스 배경색 (오류 예시이므로 붉은 계열로 변경)
    colframe=gray!50!gray, % 박스 테두리색
    fonttitle=\bfseries,   % 제목 폰트 굵게
    arc=1mm,               % 박스 모서리 둥글게
    boxrule=0.5pt,         % 테두리 두께
    fontupper=\scriptsize,
        boxsep=1pt,           % ← 전체 여백 줄이기
    left=2pt,             % ← 왼쪽 여백
    right=2pt,            % ← 오른쪽 여백
    top=2pt,              % ← 위쪽 여백
    bottom=2pt,            % ← 아래쪽 여백
]
\textbf{Dialogue 2 (Translated from Korean):}
{ % ← open group
\vspace{-4pt} 
\setlength{\leftmargini}{4pt}
\begin{quote}
(...)\\
\textbf{[B]:} (Salutes) Hey.\\
\textbf{[C]:} Hey.\\
\textbf{[B]:} Hey... What's this? A drowned body? Doesn't even look that deep to me.\\
\textbf{[A]:} Doesn't seem like he drowned.\\
\textbf{[B]:} Then what, a dumped body?\\
\textbf{[A]:} Nah... doesn't look dumped either. Go take a closer look. Go on.\\
\textbf{[B]:} Then what the hell is it?\\
\textbf{[A]:} Hey B, you know, brace yourself before you look.\\
\textbf{[B]:} You kidding me? Damn it... shit...
\end{quote}
}
\vspace{-4pt} 
\noindent\textbf{Ground Truth:} Friend/Coworker \\
\textbf{Prediction:} Supervisor-Subordinate (Qwen, Gemini, GPT)
\end{tcolorbox}

In Dialogue 2, GPT-4o identifies three cues: (1) A and B converse casually without honorifics, (2) the topic concerns work, and (3) B complies with A's instruction. The model places greatest weight on the last cue, interpreting the interaction as hierarchical and labels them as supervisor–subordinate. For Korean speakers, this is implausible, since subordinates are expected to use honorifics when addressing a superior. The absence of honorifics indicates the relationship is not hierarchical, but rather that of coworkers or friends. This shows that even when the models detect relevant cues, they are unable to prioritize and integrate them within the social context. We provide original Korean scripts for Dialogue 2 in Appendix~\ref{appendix:qual_examples_original_k_diag}.
\begin{figure}[t]
  \centering

  % --- figure* next ---
  \includegraphics[width=\columnwidth]{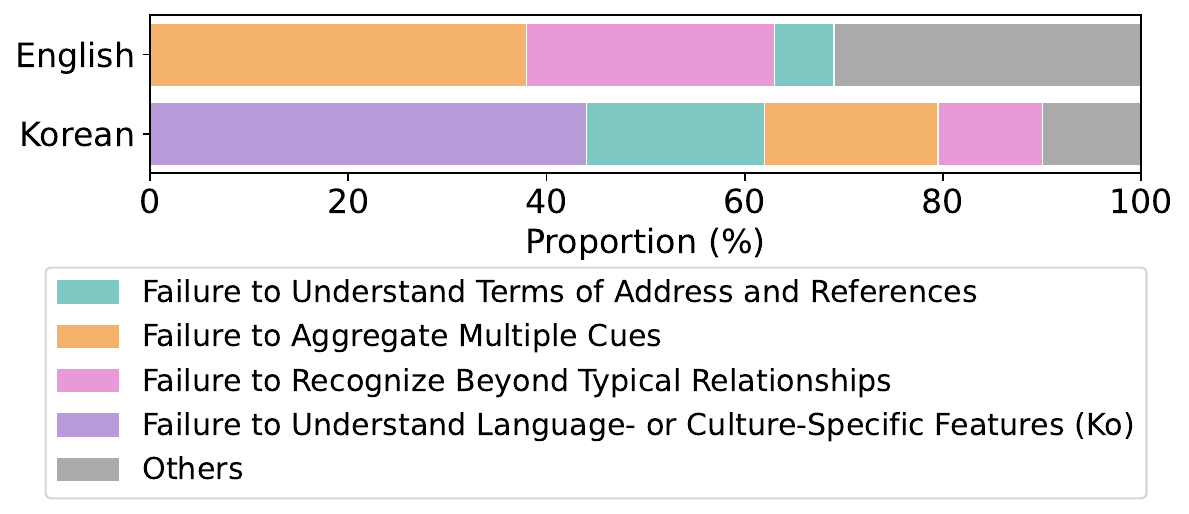}
  \caption{\textbf{Distribution of GPT-4o’s 30 failure cases by error type in English and Korean.}}
  \label{fig:breakdown_qual}
\vspace{-1mm}
\end{figure}

%Building on this limitation, we further investigate whether supplementing textual signals with visual cues can help models disambiguate relational dynamics. A small-scale exploration of this direction is provided in Appendix~\ref{appendix:qual-eval_visual_cues}\todo{}.

% These failure cases provide insights into why LLMs exhibit lower accuracy when using the CoT method in relationship inference task. The process of holistic reasoning fundamentally differs from a step-by-step reasoning approach. This highlights the need for other methods that can better handle holistic reasoning in social contexts where a task requires synthesizing conflicting or contextual clues into a cohesive conclusion

\paragraph{Failure to Recognize Atypical Relationships}
Models frequently struggle to recognize relationships that deviate from conventional or stereotypical patterns, such as non-hierarchical (equal) exchanges between parents and children or hierarchical conversations between married couples.

\begin{tcolorbox}[breakable,
    colback=gray!5!white,   % 박스 배경색 (오류 예시이므로 붉은 계열로 변경)
    colframe=gray!50!gray, % 박스 테두리색
    fonttitle=\bfseries,   % 제목 폰트 굵게
    arc=1mm,               % 박스 모서리 둥글게
    boxrule=0.5pt,         % 테두리 두께
    fontupper=\scriptsize,
        boxsep=1pt,           % ← 전체 여백 줄이기
    left=2pt,             % ← 왼쪽 여백
    right=2pt,            % ← 오른쪽 여백
    top=2pt,              % ← 위쪽 여백
    bottom=2pt            % ← 아래쪽 여백
]
\textbf{Dialogue 3 (English):}
{ % ← open group
\vspace{-4pt} 
\setlength{\leftmargini}{4pt}
\begin{quote}
\textbf{[A]:} So you're seeing Mom tomorrow, huh? At my parent-teacher thing?\\
\textbf{[B]:} Yeah.\\
\textbf{[A]:} First time in a while.\\
\textbf{[B]:} Yeah, but no biggie.\\
\textbf{[B]:} Hey, what's with the moping?\\
\textbf{[A]:} Nothing. It's just... there's this girl.\\
\textbf{[B]:} Oh yeah? You like her?\\
\textbf{[A]:} I like [C]. This girl's my soulmate. I'm like crazy, stupid, in love with her. And she wants someone else.\\
\textbf{[B]:} But she's your soulmate?\\
\textbf{[A]:} Yeah.
\end{quote}
\vspace{-4pt} 
\noindent\textbf{Ground Truth:} Parent-Children \\
\textbf{Prediction:} Siblings (Llama, GPT, Gemini)
}
\end{tcolorbox}

In Dialogue 3, all human annotators agree on the parent–child relationship, yet the models reject it: \textit{``less likely since the conversation feels more peer-like rather than hierarchical or guiding''} (GPT-4o), \textit{``the casual tone and discussion about a crush imply a more peer-like relationship''} (Qwen-3-8b), and \textit{``if [B] is the parent, they might not discuss the girl in such a casual way''} (Llama-3.1-8b-instruct). Gemini-2.5-flash likewise dismisses the parent–child relationship, reasoning that B is a bachelor and A's parents are deceased, concluding it is not a traditional parent–child relationship, revealing a stereotyped conception of family roles.

\paragraph{Failure to Understand Language- or Culture-Specific Features (Ko)}
In Korean, this error type accounts for the largest share of failures. Most confusions stem from the misinterpretation of terms of address  and honorifics. For example, unlike English, \textit{eomeoni}, literally \textit{“mother,”} can also be used to address a friend’s mother or an older woman, yet the model often predicts a parent–child relationship whenever it appears. This issue is especially pronounced in dialogues containing culturally specific terms such as kinship expressions. For instance, Qwen misinterprets \textit{Hyungsoo} (older brother's wife) as \textit{“older brother”}, and \textit{Hyungnim} as \textit{“father”}, resulting in a complete failure.
Honorifics are also frequently misinterpreted—for example, equal relationships (e.g., friendships) predicted as hierarchical, and vice versa.

Additionally, we manually examine 30 failure cases of GPT-4o (best performing model). Figure~\ref{fig:breakdown_qual} presents their distribution across error types in English and Korean. In English, the majority of errors arise from Failure to Aggregate Multiple Cues (36.7\%). In contrast, Korean errors are predominantly caused by difficulties in handling Language and/or Culture-Specific Features (46\%), with smaller proportions attributed to the other categories. This disparity highlights the model’s difficulty in identifying relational cues embedded in Korean-specific cultural and linguistic markers, such as terms of address and honorific systems, thereby revealing the culturally dependent nature of social relationship reasoning.

\section{Does Providing Additional Social Information Help?}
\label{sec:discussion-cue}
% Humans interpret social relationships considering demographic cues and relational dimensions~\cite{nguyen-etal-2016-survey}. Here, we examine whether such information can similarly improve the social reasoning abilities of LLMs.\footnote{We observe that models often leverage relational dimensions to infer the relationship; see Appendix~\ref{sec:discussion-cues-pattern} for examples.} We use four models—GPT-4o, Gemini-2.5-flash, Qwen-3-8b, and Llama-3.1-8b-instruct—selecting one model from each family, while excluding thinking models and Korean-specialized models.

% \subsection{Impact of social information on Model Performance}
% We conduct experiments to evaluate how providing social information influences model performance in inferring interpersonal relationships. We use two information types: demographic cues (age, gender) and relational dimensions (intimacy, hierarchy, formality).
Humans interpret social relationships using demographic cues and relational dimensions~\cite{nguyen-etal-2016-survey}. Our initial analysis also suggests that LLMs often rely on relational dimensions when inferring relationships; Appendix~\ref{sec:discussion-cues-pattern} provides examples. Motivated by this, we investigate whether such social information can similarly enhance the social relationship reasoning abilities of LLMs using four models—GPT-4o, Gemini-2.5-flash, Qwen-3-8B, and Llama-3.1-8B-instruct—each representing a different model family while excluding thinking-enabled and Korean-specialized models. We examine how providing social information influences performance in inferring social relationships, considering two types of information: demographic cues (age, gender) and relational dimensions (intimacy, hierarchy, formality).

\paragraph{Experimental Setting} We design six experimental settings with two variables. First, we vary the type of social information: (i) age/gender only, (ii) relational dimensions only, or (iii) both. Second, we vary the source of social information: (a) human-annotated gold data, available in the dataset as metadata (see \S~\ref{sec:method_human_annotation} for illustration), or (b) model-generated predictions, where the model infers each type of social information and incorporates these predictions into the social relationship reasoning. The accuracy of these predictions is reported in Table~\ref{tab:sub_dims_accuracy} in the appendix. Detailed experimental settings appear in Appendix~\ref{appendix:prompt_section_6}.

\begin{figure}[t]
  \centering

  % --- figure* next ---
  \includegraphics[width=\linewidth]{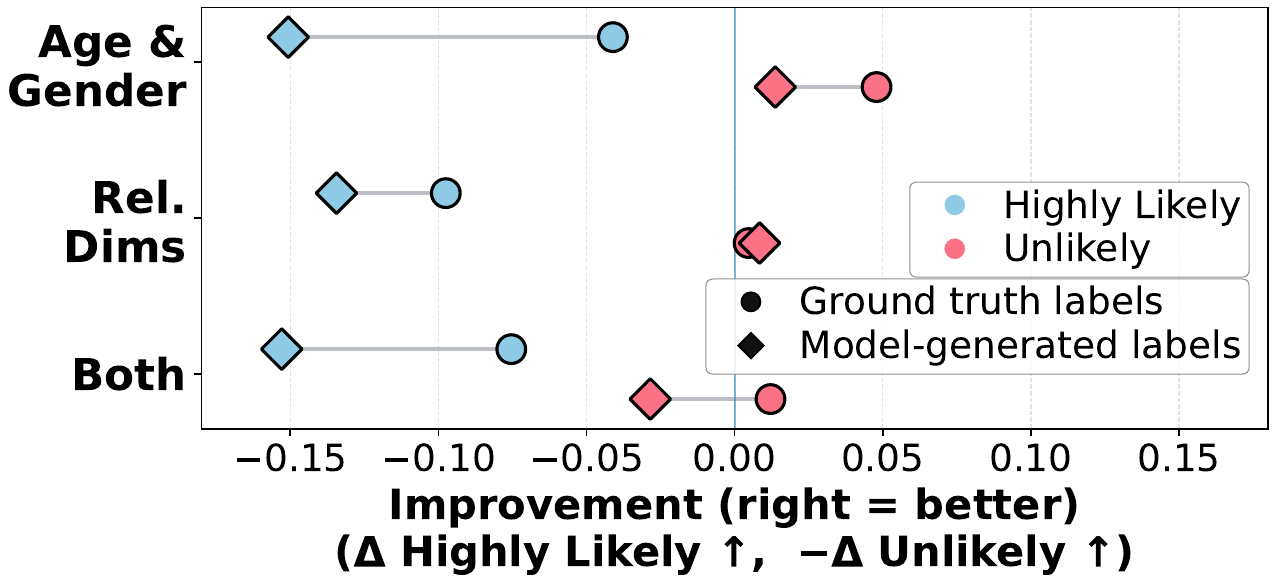}
\caption{\textbf{Impact of Relational Information on GPT-4o's Performance.}
Positive values indicate improvement, while negative values indicate deterioration after adding relational information.}
\vspace{-1mm}
\label{tab:reesults_input_subdims}

\end{figure}

\paragraph{Results With Ground Truth Labels}
Figure~\ref{tab:reesults_input_subdims} shows GPT-4o results across six settings on the English dataset. Providing human gold information yields no substantial or consistent performance gains, but it reduces the proportion of \textsc{Unlikely} predictions.  This suggests that while such information may not directly guide identification of \textsc{Highly~Likely} relationships, it helps models avoid \textsc{Unlikely} ones. The tendency holds across models, except for Qwen-3-8B. For instance, when the intimacy label for two speakers is ``Intimate,'' the model’s initial inference often shifts to more intimate relationship categories after this label is provided (e.g., Strangers $\rightarrow$ Romantic Interest, 3.3\%). Similarly, when given the label ``No hierarchy'', the most common change is also Parent–Children → Friends (2.9\%). Thus, dimension labels provide additional cues about relationships, enabling the model to incorporate them and reduce implausible predictions.
However, these changes do not always yield correct reasoning. Sometimes models over-rely on dimensional labels rather than context. For instance, in an atypical ``close'' superior–subordinate relationship, GPT-4o misinterprets the interaction due to the intimate tone, even when clear terms of address are present.

% Figure~\ref{tab:reesults_input_subdims} reports GPT-4o results across six settings on the English dataset. Providing gold human information does not yield substantial or consistent gains, but it consistently reduces \textsc{Unlikely} predictions. This suggests that such information may not directly help identify \textsc{Highly~Likely} relationships, but it helps models avoid implausible ones. This trend holds across models except Qwen-3-8B.
% Qualitatively, when an \textsc{Unlikely} prediction is revised into a more plausible one, labels often shift toward relationships consistent with the provided dimensions. For example, under the label Intimate,'' predictions sometimes move from Strangers to Romantic Interest (3.3\%). Likewise, with No hierarchy,'' a common shift is Parent--Children to Friends (2.9\%). These patterns suggest that dimension labels provide additional relational cues that models can use to reduce implausible outputs.
% However, these shifts are not always correct: models can over-rely on dimensional labels at the expense of dialogue evidence. For instance, in an atypically ``close'' superior--subordinate interaction, GPT-4o is misled by the intimate tone despite clear terms of address.

\paragraph{Results With Model-Generated Labels}
With model-generated information, auxiliary labels do not consistently help because they are often inaccurate. For example, GPT-4o achieves under 60\% accuracy on age and gender and below 75\% on relational dimensions (see Table~\ref{tab:sub_dims_accuracy} for accuracy across four models). Consistent with this, inferred social information is more accurate in cases where it improves relationship reasoning than in cases where it degrades performance (Table~\ref{tab:subdim_performance_analysis}).
\begin{table}[t]
\centering
\small
\resizebox{\linewidth}{!}{%
\begin{tabular}{lcc}
\toprule
Social. Info. & \shortstack{\textbf{Improved} \\ \textbf{(Unlikely$\rightarrow$Likely)}} & 
             \shortstack{\textbf{Deteriorated} \\ \textbf{(Likely$\rightarrow$Unlikely)}} \\
\midrule
\texttt{Age \& gender} & \textbf{72.8} & 65.5 \\
\texttt{Rel. dims.}    & \textbf{53.3} & 50.7 \\
\bottomrule
\end{tabular}
}
% \caption{\textbf{Performance comparison of improved versus deteriorated predictions across Relational Dimensions (Rel. dims.) and Age/Gender.}
% Improved cases are defined as \textsc{Unlikely} $\rightarrow$ \textsc{\{Less/Highly\}~Likely}, while deteriorated cases are defined as \textsc{\{Less/Highly\}~Likely} $\rightarrow$ \textsc{Unlikely}.}
\caption{\textbf{Accuracy of social information inference in cases where social relationship reasoning improved or deteriorated.}}
\vspace{-1mm}
\label{tab:subdim_performance_analysis}
\end{table}

These results suggest that demographic cues and relational dimensions, which humans naturally rely on, can facilitate \task. However, current LLMs are limited in their ability to infer these dimensions. Therefore, instructing LLMs to infer these factors before identifying the social relationships is ineffective. Results for other models are provided in Table~\ref{tab:results_input_subdims}-\ref{tab:results_input_subdims_ko} in the Appendix~\ref{appendix:integrating_social_info}. In table~\ref{tab:corr_subdims} of Appendix~\ref{appendix:integrating_social_info}, we examine the link between social-information inference and relationship reasoning using separate logistic regressions for each factor.

\section{Conclusion}
\label{sec:conclusion}
We introduce a bilingual dataset \dataset{} to investigate the limitations of current LLMs in \task. 
Our experiments show that most models perform suboptimally across English and Korean, and often infer \textsc{Unlikely} relationships.
Our analyses (\S\ref{sec:exp}) reveal that current reasoning techniques such as CoT do not consistently benefit social reasoning. 
Furthermore, we provide an analysis on where LLMs fail, especially focusing on cases where models respond with \textsc{Unlikely} relationships.

Our findings suggest several directions for improving models’ social relationship reasoning. At inference time, providing more explicit relational and social cues may help reduce implausible inferences. At training time, broader exposure to rare, atypical, and culturally diverse relationships may help reduce models’ reliance on common relationship patterns. More broadly, our cross-linguistic results, including improved performance of Korean-specific models on Korean dataset, demonstrate the importance of language and culture-specific approaches to advance LLMs' social reasoning abilities. We hope \dataset{} provides a useful starting point for improving LLMs' \task across diverse social contexts in different languages and cultures.

% \newpage
\section*{Limitations}
While our dataset covers both English and Korean, our analysis remains limited to these two languages and may not generalize to other cultural contexts. In addition, because our data comes from movie scripts, it may not fully reflect real-world conversation. Still, given the privacy and labeling challenges of collecting large-scale real dialogue with reliable relationship annotations, movie scripts provide a practical proxy for this task. Future work should extend the benchmark to more languages and more realistic sources, such as privacy-preserving real conversations and human-AI dialogue logs.

Additionally, as discussed in \S\ref{sec:qual-eval}, we analyze CoT traces to characterize model behavior. However, we acknowledge that these traces may reflect post-hoc rationalizations rather than the mechanisms that produced the final answer.
\section*{Ethics Statement}
This study involves human annotation on pre-existing movie scripts, which may contain harmful or offensive content due to the nature of the source material. The study was approved by KAIST IRB (KAISTIRB-2025-61), and informed consent was obtained from all participants prior to their involvement. Annotators were recruited via an institutional participant portal and compensated at hourly rates of KRW 15,000 (Korea) and USD 20 (U.S.), approximately 1.5× the local minimum wage.

\section*{Acknowledgements}
EK, JP, JO, KP, SS, and AO were supported by Institute of Information \& communications Technology Planning \& Evaluation(IITP) grant funded by the Korea government(MSIT) (No. RS-2024-00509258 and No. RS-2024-00469482, Global AI Frontier Lab). They were also supported by Artificial intelligence industrial convergence cluster development project funded by the Ministry of Science and ICT(MSIT, Korea)\&Gwangju Metropolitan City.

We used AI assistants, including ChatGPT
 for grammar editing and refinement, and Cursor
 for coding support.\footnote{\url{https://chatgpt.com/}}\footnote{\url{https://cursor.com/}}

\newpage
\bibliography{custom,anthology}

\appendix
\clearpage
% \section{Author Contribution}
% Eunsu, Junyeong, and Juhyun were the core junior authors. Eunsu led the project, coordinated annotator recruitment and management, conducted all experiments, organized the dataset, and wrote the majority of the manuscript. Junyeong contributed extensively to implementation, including developing the annotation tool, organizing and analyzing data, conducting statistical and qualitative evaluations, and writing sections on qualitative evaluation and dataset construction. Juhyun guided the overall research direction, contributed to IRB preparation, and co-wrote the manuscript, particularly the introduction, related work, and methodology sections. 

% Seyoung supported data collection and OCR processing, and also contributed to qualitative annotations. Kiwoong contributed to script filtering during dataset preparation, and participated in the majority of qualitative annotations.

% Seza, Najoung, and Alice served as senior authors. Seza and Alice contributed to the initial ideation of the project. Najoung and Alice offered critical guidance on the overall research direction. Najoung facilitated access to the English annotator pools. Seza provided high-level comments on the annotation process. Seza and Najoung also offered substantial feedback and support throughout the writing process.
\section{Dataset}
\label{appendix:dataset}

\begin{figure*}[t]
  \centering
  \includegraphics[width=0.8\textwidth]{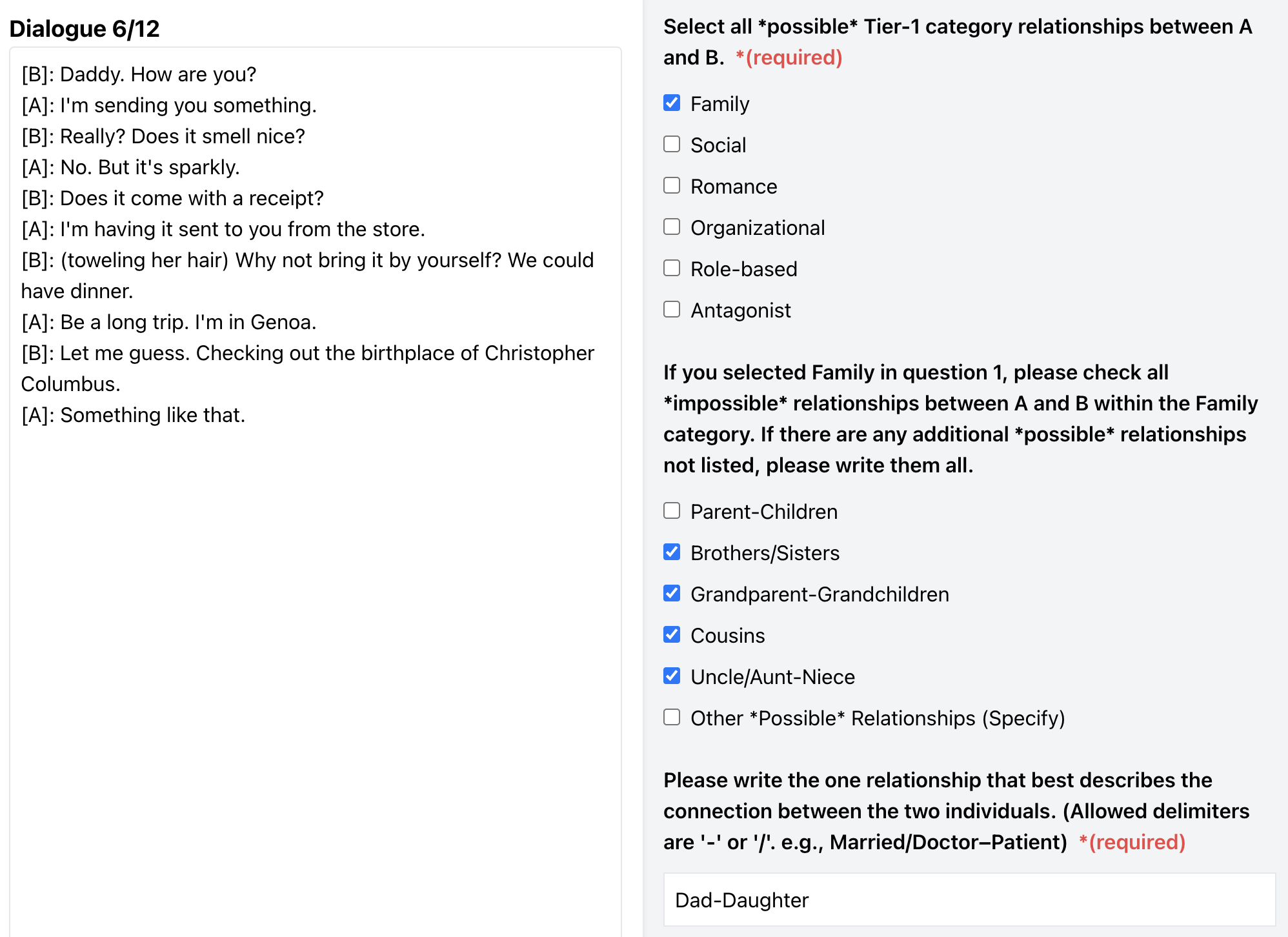}
  \caption{A screenshot of our annotation platform. The annotators can read the dialogue on the left panel and annotate the relationships and relational dimensions on the right panel.}
  \label{fig:platform}
\end{figure*}

The dialogues in \dataset{} are sourced from movie scripts. The collection process consists of the following steps: (1) movie selection, (2) raw scene collection, (3) OCR processing and human verification, (4) scene filtering, (5) anonymization, and (6) relationship annotation.

\subsection{Movie Selection}

We select the movies based on the following criteria to best capture daily real-life interactions.
Only modern-day movies released after 2000 are included, excluding medieval fantasy or alien sci-fi. To minimize exposure to violent or explicit scenes, we consider age limits, only including movies up to PG-13 for English movies and up to 15세 관람가 (suitable for audiences aged 15 and above) for Korean movies.

We collect English movie scripts from IMSDb.\footnote{https://imsdb.com} 
For Korean movies, due to the limited online dialogue resources, we visited the Korean Film Archive (KOFA)\footnote{http://www.kmdb.or.kr/} to collect physical copies of movie scripts. From KOFA, we only collect one-third of each movie script to adhere to the data use policy. Additional movie scripts were sourced from Filmmakers Online Community.\footnote{https://www.filmmakers.co.kr/}
In total, we collect\
60 movies~(28 English, 32 Korean) across various genres.

\subsection{Raw Scene Collection and Processing}

We use NAVER CLOVA OCR API\footnote{https://www.ncloud.com/product/aiService/ocr} to extract dialogue texts from physical copies of Korean movie scripts from KOFA. After OCR, we use GPT-4o to further process and clean the text into a structured format(as dialogue in Figure~\ref{fig:platform}).
Human annotators then verify and correct the outputs based on the original PDF files. As a result of this process, we obtain 16k English and 7k Korean scenes.

\subsection{Scene Filtering}

Scenes are filtered to include at least four utterances, and involve two to three speakers. To maximize speaker diversity, we prioritize scenes featuring unique character pairs within each movie.
Using these criteria, we select 1,322 scenes from an initial pool of 23k scenes, comprising 698 English and 624 Korean scenes. The selected movies and the number of scenes per movie are listed in Table~\ref{tab:movies}.

\subsection{Anonymization}

To mitigate potential data contamination (e.g., identifying the source movie and providing a response based on parametric knowledge about the movie) and reduce bias (e.g., gender inference), all character names are automatically replaced with placeholders such as \texttt{[A]} and \texttt{[B]}. Any names not covered by this process are further verified and anonymized manually.

\subsection{Human Annotation}
\label{appendix:Human Annotation}
We construct the gold label set from human annotators who have over ten years of experience in the target language and culture. Annotators include undergraduate and graduate students in South Korea and the United States, compensated at 1.5 times the local minimum hourly wage. Payment is processed via the Upwork platform.\footnote{\href{https://www.upwork.com/}{https://www.upwork.com/}} Actual annotation is conducted on our own annotation platform (Figure~\ref{fig:platform}). All annotations are conducted under the protocol approved by IRB.

\paragraph{Recruitment and Management of Human Annotators} As shwon in Figure~\ref{fig:platform}, the anonymized scene appears on the left, and annotation questions on the right. We recruite 17 English annotators (7 male, 10 female) and 14 Korean annotators (5 male, 9 female).  We obtained their informed consent.
Before starting the annotation, annotators attend an introductory Zoom session led by one of the authors covering data usage policies and guidelines.

All annotators are undergraduate or graduate students enrolled at universities in the United States or Korea. The Korean annotators are all native speakers, while the English annotators are either U.S. citizens or individuals who have lived in the United States for over 10 years. For quality control, applicants are asked to complete the task on three sample items during recruitment, and their responses are reviewed by the authors to select the final annotators. After selection, annotators participate in an orientation session and a training phase designed to support them in performing the task as effectively as possible. 

We provide the participant recruitment announcement below.
\begin{tcolorbox}[
  colframe=gray,
  colback=gray!10,
  boxrule=0.5mm,
  sharp corners,
  title={Participant Recruitment (English)},
  fonttitle=\bfseries,
  coltitle=black,
  breakable
]
\small
\setlist[itemize]{leftmargin=*, nosep}

\noindent\begin{tabularx}{\linewidth}{@{}>{\bfseries}p{0.22\linewidth} X@{}}
Overview &
We are recruiting participants for a research experiment that evaluates the conversational understanding abilities of language models. This study builds a dataset to assess models' social reasoning in dialogue. Participants will read short dialogues from movie scripts and label the social relationships between speakers. \\[0.4em]

What you will do &
\begin{itemize}
  \item Identify \textbf{social role-based relationships} (e.g., parent--child, romantic partners, mentor--mentee).
  \item Label \textbf{relationship aspects} (e.g., intimacy/closeness, hierarchy/power, purpose: work-oriented vs.\ casual).
  \item Infer \textbf{speaker attributes} (e.g., gender and approximate age).
\end{itemize} \\[0.2em]

Eligibility &
\begin{itemize}
  \item Comfortable using web-based interfaces for research participation.
  \item Fluent in English and highly familiar with U.S.\ culture (e.g., lived in the U.S.\ for 10+ years).
  \item Age 18 or older.
  \item Not offended by dialogues that may include profanity, offensive language, or depictions of violence.
  \item Registered (or able to register) as a participant on Upwork Platform.
\end{itemize} \\[0.2em]

IRB Safety Notes &
In accordance with Institutional Review Board (IRB) guidelines, we cannot recruit individuals directly supervised by the research lead, nor undergraduate students under the age of 18. Movie scripts may contain profanity, offensive language, and morally questionable situations. Participation is voluntary, and you may withdraw at any time without penalty. \\
\end{tabularx}
\end{tcolorbox}

\paragraph{Annotation Details}For social relationships, annotators are given an initial set of possible relationships (Table~\ref{tab:initial-relationships}), adapted from \citet{tigunova-etal-2021-pride}, and asked to mark \textsc{Unlikely} ones. To reduce workload, annotators first choose \textsc{Likely} relationship categories for each dialogue, then select the \textsc{Unlikely} relationships from the list of specific relationships in those categories. They provide up to five open-ended answers describing relationships that best characterize the interaction. These serve as candidates for \textsc{Likely} relationships.

For relational dimensions, we provide annotators with definitions of each dimension and ask them to rate dialogues on a 5-point scale. The dimensions we annotate are: intimacy (from strongly intimate to strongly unintimate), formality (from strongly formal to strongly informal), and hierarchy (A> >B, A>B, A=B, A<B, A< <B). When constructing the gold label set, we collapse the ratings into a 3-point scale (e.g., intimate, neutral, unintimate) and assign the majority-voted label.
The inter-annotator agreement is reported in Table~\ref{tab:agreement}. For \textsc{Highly~Likely} and \textsc{Unlikely} relationship labels, standard single-label agreement measures are not appropriate because annotators may assign multiple labels to a single dialogue. We therefore compute mean pairwise Jaccard similarity over annotators' label sets, after normalization for open-ended \textsc{Highly~Likely} responses. Higher values indicate greater overlap between annotators' selected label sets.

% Regarding the social relationship between the characters, we provide an initial set of possible relationships (Table~\ref{tab:initial-relationships}) and ask the annotators to identify the \texttt{``unlikely''} relationships. (The initial set of likely relationships is obtained from \citet{tigunova-etal-2021-pride}.) 

% To reduce annotator workload, we first ask them to select the \texttt{likely} relationship categories for each dialogue. Within those selected categories, annotators are then shown the list of specific relationships and asked to mark all relationships they consider \texttt{unlikely}. We also ask the annotators to write (up to five) open-ended answers  describing the relationships that best characterize the interaction. These open-ended answers are the candidates for \texttt{``likely''} relationships.

% Figure~\ref{fig:platform} illustrates the human annotation platform we provide to the annotators. On the left side is the anonymized scene, and on the right side are the annotation questions. All human annotations in this research project are conducted
% with the approval of IRB.

\subsection{Diversity of \dataset}
\label{appendix:dataset_diversity}
\begin{figure*}[t]
  \centering

  % --- figure* next ---
  \includegraphics[width=0.9\textwidth]{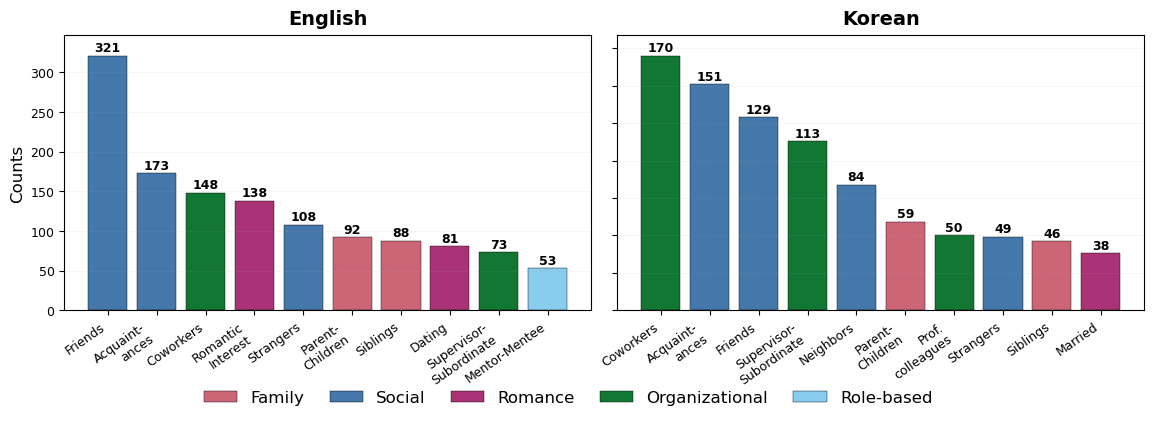}
  \caption{\textbf{Top 10 Relationships in each Language dataset.}}
  \label{fig:relation-count}

\end{figure*}

Figure~\ref{fig:relation-count} presents the ten most common relationships in each language. Both the English and Korean datasets frequently include social (e.g., friends, acquaintances), organizational (e.g., coworkers, supervisor-subordinate), and familial (e.g., parent-child, siblings) relationships.  

\paragraph{Language-specific relationships}Beyond these shared categories, we also examine which relationship types appear exclusively in Korean or English. Each dataset contains culturally specific relationships that reflect distinct social roles and lexicalizations.

Korean-only relations include \textit{North Korean soldier--citizen} (1), \textit{shaman--client} (2), \textit{shaman--assistant} (1), \textit{private tutor--student} (1), and \textit{student’s family acquaintance--tutor} (1). In addition, kinship terms are more fine-grained in Korean: for example, distinctions such as \textit{older brother--younger brother} (1) and \textit{older brother--sister-in-law} (1), whereas in English these are typically generalized under a single ``siblings'' category. 
English-only relations include roles such as \textit{father figure--child}, \textit{mother figure--child}, \textit{co-parents}, and \textit{babysitter--child}, reflecting cultural and social roles that are more explicitly lexicalized in English. 

These observations highlight how cultural context shapes the granularity and salience of social relationships represented in dialogue datasets.

\begin{figure}[t]
  \centering
  \includegraphics[width=\columnwidth]{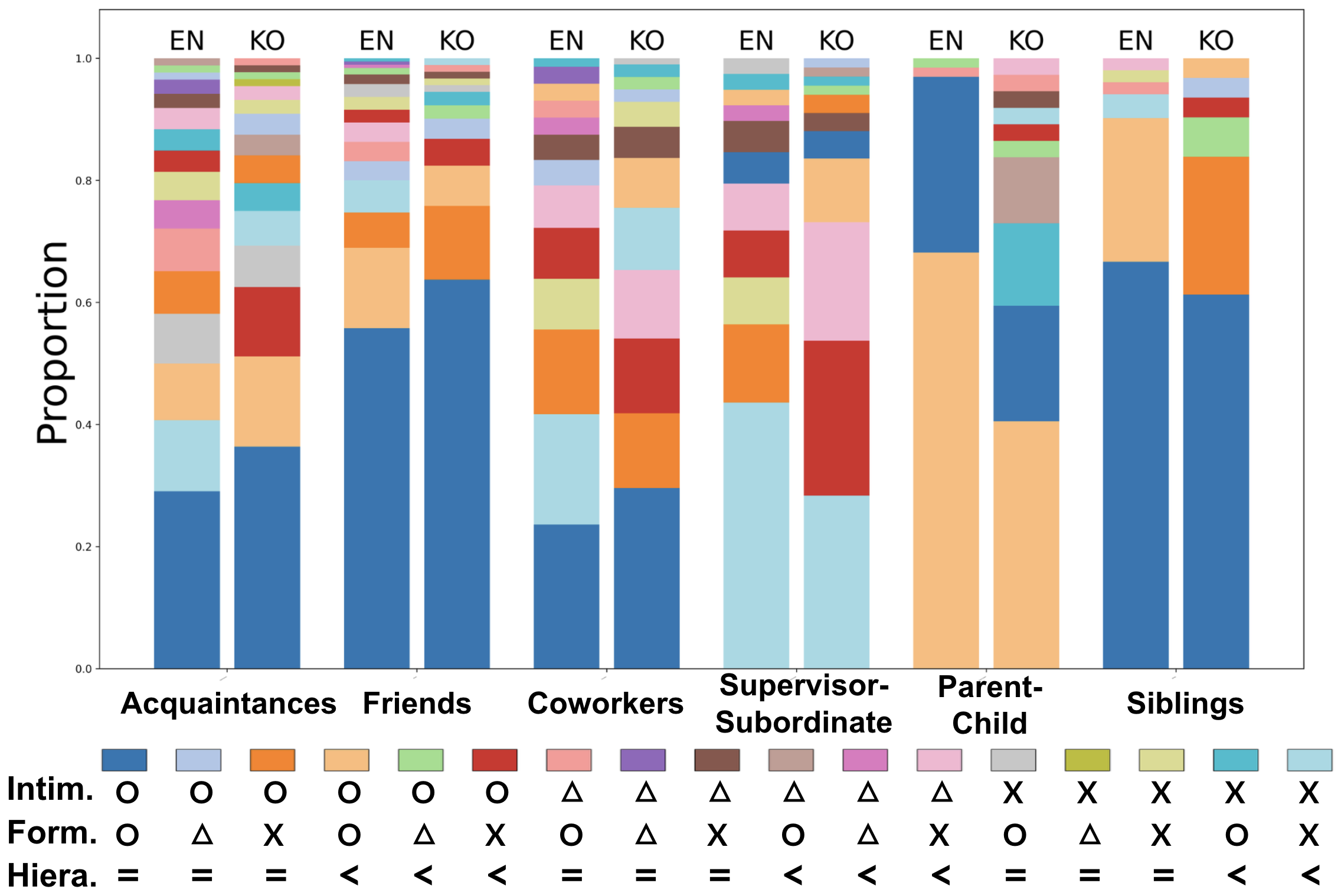}
  \caption{\textbf{Comparative analysis of relational dimension distributions for six relationship types present in both English and Korean top-10 relations.} Legend labels denote intimacy (O: intimate, X: unintimate, △: neutral), formality (O: formal, X: informal, △: neutral), and hierarchy (<: hierarchical, =: equal).}
  \label{fig:main_probable_subd_dist_en}
\end{figure}

\paragraph{Diverse interpersonal dynamics}Figure~\ref{fig:main_probable_subd_dist_en} illustrates that our dataset can capture diverse interpersonal dynamics by labeling relational dimensions. 
The typicality of certain relationships is often defined by their levels of intimacy, formality, and hierarchy~\cite{wish-perceive-1976}. For instance, friendship is generally characterized as intimate, non-hierarchical (equal), and informal. Yet, our dataset also includes atypical relationships. For instance, over 40\% of friend relationships in our dataset deviate from these typical dimensions.
\begin{table}[ht]
\centering
\resizebox{0.78\columnwidth}{!}{%
\begin{tabular}{@{}lcc@{}}
\toprule
\textbf{Annotation Type} & \textbf{EN} & \textbf{KO} \\
\midrule
Hierarchy (All) & 0.333 & 0.462 \\
Hierarchy ($2{>}$) & 0.416 & 0.550 \\
Formality (All) & 0.408 & 0.469 \\
Formality ($2{>}$) & 0.513 & 0.562 \\
Intimacy (All) & 0.314 & 0.375 \\
Intimacy ($2{>}$) & 0.426 & 0.458 \\
\textsc{Highly-Likely} (Jaccard) & 0.266 & 0.343 \\
\textsc{Unlikely} (Jaccard) & 0.837 & 0.694 \\
\bottomrule
\end{tabular}%
}
\caption{Inter-annotator agreement for auxiliary relational dimensions and relationship labels. For Hierarchy, Formality, and Intimacy, we report agreement on all annotated samples (\textit{All}) and on samples where at least two annotators selected a non-neutral label ($2{>}$). For \textsc{Highly-Likely} and \textsc{Unlikely} relationship labels, we report mean pairwise Jaccard similarity because annotators may assign multiple labels to a single dialogue.}
\label{tab:agreement}
\end{table}

\section{Experiment}
\label{appendix:exp}
\subsection{Model Configuration}
\label{appendix:exp_model_config}
We evaluate nine models: GPT-4o, o3, Gemini-2.5-Flash, Qwen-3-\{8B/14b\}~\citep{yang2025qwen3},  Llama-3.1-8B-instruct~\citep{grattafiori2024llama}, A.X-4.0-light-7B, Kanana-8B
, and Exaone-4.0-30B~\cite{LG2025exaone40}.\footnote{\href{https://platform.openai.com/docs/models/gpt-4o}{https://platform.openai.com/docs/models/gpt-4o}\\ \href{https://platform.openai.com/docs/models/o3}{https://platform.openai.com/docs/models/o3} \\ \href{https://github.com/SKT-AI/A.X-4.0}{https://github.com/SKT-AI/A.X-4.0}\\ \href{https://github.com/kakao/kanana}{https://github.com/kakao/kanana}}
For GPT-4o, we use the gpt-4o-2024-05-13 version via the OpenAI API. For Gemini-2.5-Flash, we use OpenRouter (\href{https://openrouter.ai}{openrouter.ai}). We set the temperature of the models to 1.
\subsection{Prompt}
\label{appendix:exp_prompts}
\subsubsection{LLM-as-a-Judge Prompt}
\begin{tcolorbox}[
    colframe=gray,
    colback=gray!10,
    boxrule=0.5mm,
    sharp corners,
    title={LLM-as-a-Judge Prompt},
    fonttitle=\bfseries,
    coltitle=black,breakable
]
\small 
You are a judge that evaluates the correctness of relation classification outputs from the model.  

Your task is to check each relation in the model's output and determine whether it belongs to the *Possible* or *Impossible Relation Sets* provided.

For each relation:
\begin{itemize}
    \item If it is included in the Possible Relation Sets, mark \texttt{"is\_in\_possible": 1}, otherwise mark it as 0.  
    \item Similarly, if it is included in the Impossible Relation Sets, mark \texttt{"is\_in\_impossible": 1}, otherwise mark it as 0.
\end{itemize}

Answer **only** in the following JSON format (no explanations):

\textbf{Possible Relation Sets:} \texttt{\{gt\_list\}}  \\
\textbf{Impossible Relation Sets:}  \\ \texttt{\{impossible\_gt\_list\}}  \\
\textbf{Model output:}  
\texttt{\{model\_outputs\}}

\textbf{Output in JSON format:}
\begin{verbatim}
{
   relation1 : {
     "is_in_possible": 1,
     "is_in_impossible": 0
   },
   relation2 : {
     "is_in_possible": 0,
     "is_in_impossible": 1
   },
   ...
}
\end{verbatim}
\textbf{Output:}
\end{tcolorbox}
\subsubsection{Inference Prompt}
\begin{tcolorbox}[
    colframe=gray,
    colback=gray!10,
    boxrule=0.5mm,
    sharp corners,
    title={Prompt Template},
    fonttitle=\bfseries,
    coltitle=black,title=Inference Prompt (EN),breakable
]
\small 
Read the following conversation and guess the relationship of the participants [A] and [B].  
When guessing the relationship, refer to the following examples of relationships:

\texttt{\{example\_relations\}}

If the relationship matches one of the examples above, use it as is,  
but if the relationship does not fit any of the examples, describe the relationship yourself.  

Your answer about the relationship must be in JSON format:
\begin{verbatim}
{
  "relation": ""
}
\end{verbatim}

\texttt{\{Additional Information\}}\\
\textbf{Conversation:}  
\texttt{\{dialogue\}}

\textbf{Output (JSON):}
\end{tcolorbox}
\begin{tcolorbox}[
    colframe=gray,
    colback=gray!10,
    boxrule=0.5mm,
    sharp corners,
    title={Prompt Template},
    fonttitle=\bfseries,
    coltitle=black,title=Inference Prompt (Ko),breakable
]
\small 
다음 대화를 읽고, 등장인물 A와 B의 관계를 추측하시오.  

관계를 추측할 때는 다음의 관계 예시를 참고하시오:

\texttt{\{example\_relations\}}

만약 위의 예시에 해당하는 관계라면 그대로 사용하고,  
예시에 없는 관계라고 판단되면 해당 관계를 직접 서술하시오.  

관계에 대한 최종 답은 반드시 JSON 형식으로 답변하시오:
\begin{verbatim}
{
  "relation": ""
}
\end{verbatim}
\texttt{\{Additional Information\}}\\
\textbf{대화:}  
\texttt{\{dialogue\}}

\textbf{Output (JSON):}
\end{tcolorbox}

\begin{tcolorbox}[
    colframe=gray,
    colback=gray!10,
    boxrule=0.5mm,
    sharp corners,
    title={Example Relations},
    fonttitle=\bfseries,
    coltitle=black,
    breakable
]
\small 
\textbf{English Relations}
\begin{verbatim}
[
  "Parent-Children",
  "Brothers/Sisters",
  "Grandparent-Grandchildren",
  "Cousins",
  "Uncle/Aunt-Niece",
  "Friends",
  "Acquaintances",
  "Neighbors",
  "Strangers",
  "Romantic Interest",
  "Dating",
  "Married",
  "Engaged",
  "Friends with benefits",
  "Affair",
  "Ex-relationship",
  "Coworkers",
  "Professional colleagues",
  "Supervisor-Subordinate",
  "Mentor-Mentee",
  "Teacher-Student",
  "Lawyer-Client",
  "Doctor-Patient",
  "Landlord-Tenant",
  "Competitive relationship",
  "Rivalry",
  "Arch-enemies"
]
\end{verbatim}

\textbf{Korean Relations}
\begin{verbatim}
[
  "부모-자식",
  "형제/자매/남매",
  "조부모-손주",
  "사촌",
  "삼촌/이모/고모-조카",
  "단짝 친구",
  "친구",
  "지인",
  "이웃",
  "모르는 사이",
  "썸",
  "연애",
  "부부",
  "약혼관계",
  "Friends with benefits",
  "불륜관계",
  "전애인 관계",
  "동료",
  "직장 동료",
  "상관-부하직원 관계",
  "멘토-멘티",
  "선생-제자",
  "변호사-고객",
  "의사-환자",
  "집주인-세입자",
  "경쟁관계",
  "라이벌 관계",
  "숙적"
]
\end{verbatim}
\end{tcolorbox}

\textbf{CoT setting:}  
We append \textit{``Think step by step''} at the end of the prompt to encourage chain-of-thought reasoning.

\subsubsection{Prompts Used in \S\ref{sec:discussion-cue}}
\label{appendix:prompt_section_6}
For the experiment in \S\ref{sec:discussion-cue}, we add additional information to the prompt. The additional information consists of two types: Age \& Gender and Relational Dimensions. The prompts for each type are as follows.

In the Ground Truth Labels setting, we fill the placeholders \{age\_gender\_info\} and \{relational\_dimensions\_info\} with human-annotated gold labels. In the Model-Generated Labels setting, the model is first asked to separately infer each type of information, and the inferred information is then inserted back into the corresponding placeholders.
\begin{tcolorbox}[
    colframe=gray,
    colback=gray!10,
    boxrule=0.5mm,
    sharp corners,
    title={Additional Information (En)},
    fonttitle=\bfseries,
    coltitle=black,breakable
]
\small 
\textbf{Base setting:} None

\medskip
\textbf{Age \& Gender:}  
The age and gender information of the participants [A] and [B] are as follows. Please refer to them when inferring the nature of their relationship.  
\texttt{\{age\_gender\_info\}}

\medskip
\textbf{Relational Dimensions:}  
The Intimacy level, Pleasure level, and Hierarchy level between A and B in the conversation are as follows. Please refer to them when inferring the nature of their relationship.  
\texttt{\{Relational Dimensions\_info\}}

\medskip
\end{tcolorbox}

\begin{tcolorbox}[
    colframe=gray,
    colback=gray!10,
    boxrule=0.5mm,
    sharp corners,
    title={Additional Information (Ko)},
    fonttitle=\bfseries,
    coltitle=black,breakable
]
\small 
\textbf{Base setting:} None

\medskip
\textbf{Age \& Gender:}  
등장인물 A와 B의 나이와 성별은 다음과 같다. 그들의 관계의 성격을 추론할 때 참고하라.
\texttt{\{age\_gender\_info\}}

\medskip
\textbf{Relational Dimensions:}  
대화에서 A와 B 사이의 친밀감(Intimacy) 수준, 격식(Formality) 수준, 그리고 위계(Hierarchy) 수준은 다음과 같다. 그들의 관계의 성격을 추론할 때 참고하라.
\\
\texttt{\{Relational Dimensions\_info\}}

\medskip
\end{tcolorbox}
\subsection{Validating LLM-as-a-Judge}
\label{appendix_exp_validate_judge}
To validate the accuracy of GPT-4o as an evaluator, we sample 100 question–answer pairs for each language, and two authors independently verify the results. This verification process shows that GPT-4o correctly evaluates {97.85\%} of the responses in English and {86.2\%} in Korean, with its judgments matching those of two authors on 96\% of the data points (Cohen’s $\kappa$ = 0.58).
\subsection{Results}
\label{appendix:exp-result}
\begin{table*}[ht!]
\centering
\small
\begin{tabular}{lcccc}
\toprule
\multirow{2}{*}{Model} & \multicolumn{2}{c}{English} & \multicolumn{2}{c}{Korean} \\
\cmidrule(lr){2-3} \cmidrule(lr){4-5}
 & \hlhigh{\textsc{Highly~Likely} (↑)} & \ulhigh{\textsc{UnLikely} (↓)}  &  \hlhigh{\textsc{Highly~Likely} (↑)} & \ulhigh{\textsc{UnLikely} (↓)}  \\
\midrule
GPT-4o           & 0.791 & 0.109 & 0.693 & 0.219 \\
Gemini-2.5-flash & 0.758 & 0.155 & 0.589 & 0.320 \\
Qwen-3-8b        & 0.565 & 0.236 & 0.423 & 0.335 \\
Llama-3.1-8b     & 0.413 & 0.307 & 0.321 & 0.522 \\
\midrule
GPT-4.1             & 0.819 & 0.084 & 0.783 & 0.241 \\
GPT-5               & 0.794 & 0.076 & 0.790 & 0.243 \\

Qwen3-32B              &  0.562 & 0.228  & 0.552 & 0.372 \\
Llama3.3-70B-Instruct  & 0.525 & 0.154 & 0.540 & 0.381 \\
kanana2-30b-a3b-instruct & 0.556 & 0.258 & 0.561 & 0.392 \\
\bottomrule
\end{tabular}
\caption{Comparison of model performance in English (En) and Korean (Ko) datasets.  \textsc{Highly~Likely} represents the accuracy of the model's majority response being a highly likely response, while \textsc{Unlikely} indicates the error rate where the model generate an unlikely response.}

\label{tab:main_results}
\end{table*}
See Table~\ref{tab:main_results} for the main results, Table~\ref{tab:model Cot} for results with CoT prompting, and Table~\ref{tab:k_model_full} for results from Korean-specialized models.

% \paragraph{Main Results}See Table~\ref{tab:main_results} for full results.

% \paragraph{Model Performance with CoT Prompting}
\begin{table*}[ht!]
\centering
\small
\begin{tabular}{lcccc}
\toprule
\multirow{2}{*}{Model} & \multicolumn{2}{c}{En} & \multicolumn{2}{c}{Ko} \\
\cmidrule(lr){2-3} \cmidrule(lr){4-5}
 & \hlhigh{\textsc{Highly~Likely} (↑)} & \ulhigh{\textsc{UnLikely} (↓)}  &  \hlhigh{\textsc{Highly~Likely} (↑)} & \ulhigh{\textsc{UnLikely} (↓)}  \\
\midrule
GPT-4o           & 0.802 (0.011) & 0.097 (-0.012) & 0.695 (0.002) & 0.203 (-0.016) \\
Gemini-2.5-flash & 0.741 (-0.017) & 0.127 (-0.029) & 0.603 (0.014) & 0.201 (-0.119) \\
Qwen-3-8b        & 0.688 (0.133) & 0.151 (-0.126) & 0.497 (0.023) & 0.365 (-0.026) \\
Llama-3.1-8b     & 0.541 (0.087) & 0.267 (-0.041) & 0.300 (-0.058) & 0.577 (0.031) \\
\midrule
Qwen-3-32B              & 0.694 (0.132) & 0.174 (-0.054) & 0.626 (0.074) & 0.658 (0.286) \\
Llama-3.3-70B-Instruct  & 0.620 (0.095) & 0.169 (0.015) & 0.603 (0.063) & 0.536 (0.155) \\
kanana2-30b-a3b-instruct & 0.569 (0.013) & 0.235 (-0.023) & 0.601 (0.040) & 0.614 (0.222) \\
\bottomrule
\end{tabular}
\caption{Comparison of model performance with Chain of Thought Prompting across English (En) and Korean (Ko) with deltas in parentheses.}
\label{tab:model Cot}
\end{table*}

% See Table~\ref{tab:model Cot}.
% \paragraph{Performance of Korean-specialized Models}
\label{appendix:result_kmodels}
\begin{table*}[ht!]
\centering
\small
\begin{tabular}{lcccc}
\toprule
\multirow{2}{*}{Model} & \multicolumn{2}{c}{En} & \multicolumn{2}{c}{Ko} \\
\cmidrule(lr){2-3} \cmidrule(lr){4-5}
 & \hlhigh{\textsc{Highly~Likely} (↑)} & \ulhigh{\textsc{UnLikely} (↓)}  &  \hlhigh{\textsc{Highly~Likely} (↑)} & \ulhigh{\textsc{UnLikely} (↓)}  \\
\midrule
ax-4.0-light&	0.5889	&0.1934& 0.4674&	0.4127 \\
exaone-4.0-32b&	0.3178	&0.3074& 0.4092	&0.4674\\
kanana-1.5-8b	&0.4059&	0.2884& 0.328&	0.3739\\

\bottomrule
\end{tabular}
\caption{Performance of Korean Specialized models in English and Korean.}
\label{tab:k_model_full}
\end{table*}

% See Table~\ref{tab:k_model_full}.

% \subsection{Entropy among Model responses}
% \label{appendix:main_ko}
% Along with the majority response results, we examine diversity of responses across all four model responses per dialogue (Figure~\ref{fig:ko_main_add}). 

\section{Qualitative Analysis - Cues}
\label{appendix:qual_examples}
\subsection{Original Korean Dialogue}
\label{appendix:qual_examples_original_k_diag}
\begin{tcolorbox}[breakable,
    colback=gray!5!white,   % 박스 배경색 (오류 예시이므로 붉은 계열로 변경)
    colframe=gray!50!gray, % 박스 테두리색
    fonttitle=\bfseries,   % 제목 폰트 굵게
    arc=1mm,               % 박스 모서리 둥글게
    boxrule=0.5pt,         % 테두리 두께
    fontupper=\scriptsize,
        boxsep=1pt,           % ← 전체 여백 줄이기
    left=2pt,             % ← 왼쪽 여백
    right=2pt,            % ← 오른쪽 여백
    top=2pt,              % ← 위쪽 여백
    bottom=2pt,            % ← 아래쪽 여백
]

\textbf{Dialogue 2 (Korean):}
{ % ← open group
\vspace{-4pt} 
\setlength{\leftmargini}{4pt}
\begin{quote}
(...)\\
\textbf{[B]:} (경례) 어이. \\
\textbf{[C]:} 왔냐? \\
\textbf{[B]:} 야. 뭐냐? 이 익사야? 별로 깊어 보이지도 않는데.\\
\textbf{[A]:} 익사는 아닌 것 같고.\\
\textbf{[B]:} 그럼 뭐 유기?\\
\textbf{[A]:} 아하....유기도 아닌 것 같은데. 너가 가서 한번 봐봐. 한번.\\
\textbf{[B]:} 그럼 뭐야?\\
\textbf{[A]:} 야. B아. 그 마음 단단히 먹고 봐.\\
\textbf{[B]:} 장난하나. 에이씨. 자.. 에이씨.
\end{quote}
}
\vspace{-4pt} 
\end{tcolorbox}
\subsection{What cues do LLMs rely on in social relationship reasoning?}
\label{sec:discussion-cues-pattern}
To understand how models use and integrate cues to infer social relationships, we conduct a qualitative analysis on their CoT reasoning.

\paragraph{Terms of Address and Reference}

LLMs frequently leverage terms of address and references as explicit cues to infer social relationships. For instance, when a speaker use terms like \textit{``Daddy''} or \textit{``Professor [B]''}, the models infer family-based or professional relationship. Self-reference also provide valuable information. For example, a speaker introducing themselves as \textit{``Doctor [A]''} signals their professional identity as a medical practitioner, leading to LLMs suggesting relationships such as Doctor-Patient or Doctor-Doctor. Furthermore, LLMs analyze how individuals refer to third parties to understand the relationship between the referring individuals themselves. For example, if both A and B refer to a third person as \textit{``Sergeant [C]''}, the LLM can infer that A and B are likely colleagues within a military context, and that their shared use of a formal title suggests a potentially task-oriented conversation.

\paragraph{Conversation context and background}

LLMs also take into account the context of the conversation (e.g.,  \textit{a school, church, workplace, home}). They then utilize this background information to infer the social relationship or level of intimacy between the individuals involved in the dialogue.

\paragraph{Tone or Atmosphere}

LLMs also assess the emotional tone of individuals in a dialogue to judge their social dimensions, particularly intimacy and formality, utilizing that information to infer their social relationship. The models often associate \textit{casual, friendly, teasing, empathetic, or supportive} tones with more intimate relationships while \textit{aggressive, frustrated, or angry} expressions are linked to less intimate or strained relationships. Similarly, emotional expressions, whether friendly or hostile, are often connected to informal relationships while the models associate \textit{serious, indifferent, dismissive, or emotionally neutral} expressions with formal relationships.

\paragraph{Relational Dimensions}

When inferring social relationships, models often consider relational dimensions (\textit{intimacy, hierarchy, formality}) in their rationale. For instance, in a dialogue where A playfully jokes with B while B shares personal concerns, the model infers strong intimacy and suggests a close tie such as friendship or siblinghood.

However, it is important to note that while using social dimensions as cue, particularly hierarchy, LLMs often reveal social stereotypes, attributing \textit{``typical''} relational dimensions to certain relationships. For example, models assume that a parent-child inherently shares a \textit{hierarchical} relationship while a married couple would generally have a \textit{non-hierarchical (equal)} relationship. This leads to reasoning failures when the actual interaction deviates from these norms.

\section{Does Providing Additional Relational Dimension Help?}
\label{appendix:integrating_social_info}
This section provides supplementary material for Section~\ref{sec:discussion-cue}.
Tables~\ref{tab:results_input_subdims}–\ref{tab:results_input_subdims_ko} present the results across models, and Table~\ref{tab:sub_dims_accuracy} reports the accuracy of inferring relational dimension.

\paragraph{Associations Between relational dimension and \taskfull Performance}
To further examine the link between relational dimension inference and relationship reasoning, we run separate logistic regressions for each factor. Table~\ref{tab:corr_subdims} shows that most factors are positively associated with \task. This suggests that models performing well on age, gender, and relational dimension inferences also tend to perform better on overall \task, highlighting the interconnections among these dimensions.

\begin{table*}[htbp]
\centering
\small
\begin{subtable}[t]{0.46\textwidth}
\centering
\begin{tabular}{lcc}
\toprule
\textbf{Model} & \textbf{\hlhigh{$\Delta$  {Highly~Likely} (↑)}} & \textbf{\ulhigh{$\Delta$ {UnLikely} (↓)}} \\

\midrule
\textbf{GPT-4o} && \\
\hspace*{0.8em}\texttt{Age \& Gender}   & {-0.0411} & {-0.0479} \\
\hspace*{0.8em}\texttt{Rel. Dims}       & {-0.0975} & {-0.0047} \\
\hspace*{0.8em}\texttt{Both}            & {-0.0754} & {-0.0121} \\
\midrule
\textbf{Gemini-2.5-flash} && \\
\hspace*{0.8em}\texttt{Age \& Gender}   & {-0.0411} & {-0.0343} \\
\hspace*{0.8em}\texttt{Rel. Dims}       & {-0.1563} & {-0.0570} \\
\hspace*{0.8em}\texttt{Both}            & {-0.1452} & {-0.0534} \\
\midrule
\textbf{Qwen-3-8b} && \\
\hspace*{0.8em}\texttt{Age \& Gender}   & {0.0127}  & {0.0392} \\
\hspace*{0.8em}\texttt{Rel. Dims}       & {-0.1580} & {0.1209} \\
\hspace*{0.8em}\texttt{Both}            & {-0.1138} & {0.0398} \\
\midrule
\textbf{Llama-3.1-8b} && \\
\hspace*{0.8em}\texttt{Age \& Gender}   & {0.0205}  & {-0.1027} \\
\hspace*{0.8em}\texttt{Rel. Dims}       & {-0.0123} & {-0.0735} \\
\hspace*{0.8em}\texttt{Both}            & {0.0505}  & {-0.0994} \\
\bottomrule
\end{tabular}
\caption{With Ground Truth Labels}
\label{tab:eng_subdims_gt}
\end{subtable}\hfill
\begin{subtable}[t]{0.46\textwidth}
\centering
\begin{tabular}{lcc}
\toprule
\textbf{Model} & \textbf{\hlhigh{$\Delta$  {Highly~Likely} (↑)}} & \textbf{\ulhigh{$\Delta$ {UnLikely} (↓)}} \\
\midrule
\textbf{GPT} && \\
\hspace*{0.8em}\texttt{Age \& Gender}   & -0.1507 & -0.0137 \\
\hspace*{0.8em}\texttt{Sub Dims}        & -0.1344 & -0.0084 \\
\hspace*{0.8em}\texttt{Both}            & -0.1529 &  0.0285 \\
\midrule
\textbf{Gemini} && \\
\hspace*{0.8em}\texttt{Age \& Gender}   & -0.1027 & -0.0137 \\
\hspace*{0.8em}\texttt{Sub Dims}        & -0.1194 & -0.0460 \\
\hspace*{0.8em}\texttt{Both}            & -0.1711 & -0.0017 \\
\midrule
\textbf{Qwen} && \\
\hspace*{0.8em}\texttt{Age \& Gender}   & -0.0395 &  0.0321 \\
\hspace*{0.8em}\texttt{Sub Dims}        & -0.0237 &  0.0442 \\
\hspace*{0.8em}\texttt{Both}            & -0.0916 &  0.0435 \\
\midrule
\textbf{Llama} && \\
\hspace*{0.8em}\texttt{Age \& Gender}   &  0.0548 & -0.0137 \\
\hspace*{0.8em}\texttt{Sub Dims}        &  0.0099 & -0.0514 \\
\hspace*{0.8em}\texttt{Both}            & -0.0492 & -0.0145 \\
\bottomrule
\end{tabular}
\caption{With Model-Generated Labels}
\label{tab:eng_subdims_model}
\end{subtable}
\caption{Impact of Relational Information on Model Performance (English).}
\label{tab:results_input_subdims}
\end{table*}

% 색칠용 명령 정의 (이름이 없던 부분 수정)
\newcommand{\highlightpos}[1]{\ifdim #1 pt > 0pt \cellcolor{red!30}#1 \else \cellcolor{white!30}#1 \fi}

\begin{table*}[htbp]
\centering
\small
\begin{subtable}[t]{0.46\textwidth}
\centering
\begin{tabular}{lcc}
\toprule
\textbf{Model} & \textbf{\hlhigh{$\Delta$  {Highly~Likely} (↑)}} & \textbf{\ulhigh{$\Delta$ {UnLikely} (↓)}} \\
\midrule
\textbf{GPT-4o} && \\
\hspace*{0.8em}\texttt{Age \& Gender}   & {-0.0122} & {0.0041} \\
\hspace*{0.8em}\texttt{Sub Relation}    & {0.0356}  & {-0.0328} \\
\hspace*{0.8em}\texttt{Both}            & {0.0383}  & {-0.0191} \\
\midrule
\textbf{Gemini-2.5-flash} && \\
\hspace*{0.8em}\texttt{Age \& Gender}   & {0.0123}  & {-0.0285} \\
\hspace*{0.8em}\texttt{Sub Relation}    & {0.0795}  & {-0.0740} \\
\hspace*{0.8em}\texttt{Both}            & {0.0932}  & {-0.0904} \\
\midrule
\textbf{Qwen-3-8b} && \\
\hspace*{0.8em}\texttt{Age \& Gender}   & {-0.1021} & {0.1796} \\
\hspace*{0.8em}\texttt{Sub Relation}    & {-0.0740} & {0.0795} \\
\hspace*{0.8em}\texttt{Both}            & {-0.0631} & {0.0384} \\
\midrule
\textbf{Llama-3.1-8b} && \\
\hspace*{0.8em}\texttt{Age \& Gender}   & {-0.0285} & {-0.0735} \\
\hspace*{0.8em}\texttt{Sub Relation}    & {-0.0274} & {-0.1644} \\
\hspace*{0.8em}\texttt{Both}            & {0.0082}  & {-0.1315} \\
\bottomrule
\end{tabular}
\caption{With Ground Truth Labels}
\label{tab:ko_subdims_gt}
\end{subtable}\hfill
\begin{subtable}[t]{0.46\textwidth}
\centering
\begin{tabular}{lcc}
\toprule
\textbf{Model} & \textbf{\hlhigh{$\Delta$  {Highly~Likely} (↑)}} & \textbf{\ulhigh{$\Delta$ {UnLikely} (↓)}} \\
\midrule
\textbf{GPT-4o} && \\
\hspace*{0.8em}\texttt{Age \& Gender}   & {-0.0285} & {0.0490} \\
\hspace*{0.8em}\texttt{Sub Relation}    & {0.0329}  & {0.0028} \\
\hspace*{0.8em}\texttt{Both}            & {-0.0055} & {0.0165} \\
\midrule
\textbf{Gemini-2.5-flash} && \\
\hspace*{0.8em}\texttt{Age \& Gender}   & {0.0123}  & {-0.0244} \\
\hspace*{0.8em}\texttt{Sub Relation}    & {-0.0466} & {0.0795} \\
\hspace*{0.8em}\texttt{Both}            & {0.0357}  & {-0.0137} \\
\midrule
\textbf{Qwen-3-8b} && \\
\hspace*{0.8em}\texttt{Age \& Gender}   & {-0.1796} & {0.1877} \\
\hspace*{0.8em}\texttt{Sub Relation}    & {-0.0055} & {0.0329} \\
\hspace*{0.8em}\texttt{Both}            & {-0.0905} & {0.1014} \\
\midrule
\textbf{Llama-3.1-8b} && \\
\hspace*{0.8em}\texttt{Age \& Gender}   & {-0.1061} & {-0.0245} \\
\hspace*{0.8em}\texttt{Sub Relation}    & {-0.0932} & {0.1315} \\
\hspace*{0.8em}\texttt{Both}            & {-0.0165} & {-0.1342} \\
\bottomrule
\end{tabular}
\caption{With Model-Generated Labels}
\label{tab:ko_subdims_model}
\end{subtable}
\caption{Impact of Relational Information on Model Performance (Korean).}
\label{tab:results_input_subdims_ko}
\end{table*}

\begin{table*}
\centering
\footnotesize
\begin{tabular}{lcccccc}
\toprule
Model & Age & Gender & Intimacy & Formality & Hierarchy & Overall \\
\midrule
GPT-4o           & 49.1\%  & \textbf{57.85\%} & 62.7\% & 73.8\% & \textbf{71.5\%} & \textbf{60.1\%} \\
Gemini-2.5-Flash & \textbf{51.65\%} & 41.45\% & \textbf{78.3\%} & \textbf{76.8\%} & 69.9\% & 57.5\% \\
Qwen3-8b         & 30.25\% & 44.45\% & 42.5\% & 40.0\% & 54.4\% & 40.9\% \\
Llama-3.1-8b     & 42.7\%  & 36.2\%  & 61.6\% & 71.1\% & 47.4\% & 47.8\% \\
\bottomrule
\end{tabular}
\caption{Accuracy of Inferring Relational Information.}
\label{tab:sub_dims_accuracy}
\end{table*}

% 연속적인 값 컬러 스케일 (예: 파랑-빨강)
\newcommand{\posval}[1]{\cellcolor{blue!#1!white}#1}
\newcommand{\negval}[1]{\cellcolor{red!#1!white}#1}

\begin{table}[t]
\centering
\resizebox{0.9\linewidth}{!}{
\begin{tabular}{lcccc}
\toprule
Rel. Info.  & Gemini2.5 & GPT4o & Llama3.1 & Qwen3 \\
\midrule
\texttt{Age}        & \cellcolor{blue!20}0.033  & \cellcolor{blue!10}0.001  & \cellcolor{red!15}-0.051 & \cellcolor{red!30}-0.079 \\
\texttt{Gender}     & \cellcolor{red!10}-0.023 & \cellcolor{blue!40}0.116  & \cellcolor{blue!25}0.074 & \cellcolor{blue!30}0.085 \\ \midrule 
\texttt{Intimacy}   & \cellcolor{blue!15}0.044 &   \cellcolor{blue!30}0.095  & \cellcolor{blue!50}0.134 & \cellcolor{blue!20}0.070 \\
\texttt{Formality}  & \cellcolor{blue!35}0.109 & \cellcolor{blue!25}0.080  & \cellcolor{blue!30}0.099 & \cellcolor{blue!10}0.029 \\
\texttt{Hierarchy}  & \cellcolor{blue!80}\textbf{0.217} & \cellcolor{blue!60}0.164  & \cellcolor{red!5}-0.012 & \cellcolor{blue!50}0.150 \\
\bottomrule
\end{tabular}
}
\caption{\textbf{Regression coefficients for relational dimension inference and social relationship reasoning performance}}
\label{tab:corr_subdims}
\end{table}

\begin{table*}[ht]
\centering
\small
\resizebox{\textwidth}{!}{%
\begin{tabular}{@{}llllr r@{}}
\toprule
\textbf{Language} & \textbf{Movie Title} & \textbf{Movie Title (Ko)} & \textbf{Genre} & \textbf{Year} & \textbf{\# of Scenes} \\
\midrule
\multirow{29}{*}{EN} 
 % & {A Beautiful Mind}            & - & Drama, Biography, Mystery & 2001 &  \\
 & {Amelia}                      & - & Adventure, Biography, Drama & 2009 & 17 \\
 & {Autumn in New York}          & - & Drama, Romance & 2000 & 26 \\
 & {Big Fish}                    & - & Adventure, Epic, Drama & 2003 & 27 \\
 & {Bruce Almighty}              & - & Comedy, Fantasy & 2003 & 32 \\
 & {Crazy Love}                  & - & Documentary, Romance & 2007 & 21 \\
 & {Crazy, Stupid, Love.}        & - & Romance, Comedy, Drama & 2011 & 39 \\
 & {Date Night}                  & - & Romance, Comedy, Crime & 2010 & 24 \\
 & {Easy A}                      & - & Comedy, Drama, Romance & 2010 & 45 \\
 & {He's Just Not That Into You} & - & Romance, Comedy, Drama & 2009 & 16 \\
 & {Larry Crowne}                & - & Comedy, Drama, Romance & 2011 & 15 \\
 & {Monte Carlo}                 & - & Adventure, Comedy, Family & 2011 & 6 \\
 & {Moonrise Kingdom}            & - & Romance, Adventure, Comedy & 2012 & 6 \\
 & {New York Minute}             & - & Comedy, Adventure, Crime & 2004 & 59 \\
 & {Something's Gotta Give}      & - & Comedy, Drama, Romance & 2003 & 21 \\
 & {Speed Racer}                 & - & Action, Adventure, Comedy & 2008 & 7 \\
 & {The Blind Side}              & - & Drama, Biography, Sport & 2009 & 16 \\
 & {The Bounty Hunter}           & - & Comedy, Action, Romance & 2010 & 21 \\
 & {The Brothers Bloom}          & - & Comedy, Action, Adventure & 2008 & 15 \\
 & {The Curious Case of Benjamin Button} & - & Drama, Fantasy, Romance & 2008 & 22 \\
 & {The Fault in Our Stars}      & - & Drama, Romance & 2014 & 17 \\
 & {The Italian Job}             & - & Action, Crime, Thriller & 2003 & 22 \\
 & {The Invention of Lying}      & - & Comedy, Fantasy, Romance & 2009 & 20 \\
 & {The Next Three Days}         & - & Thriller, Action, Drama & 2010 & 12 \\
 & {The Pacifier}                & - & Action, Comedy, Drama & 2005 & 16 \\
 & {The Secret Life of Walter Mitty} & - & Adventure, Comedy, Romance & 2013 & 12 \\
 & {The Theory of Everything}    & - & Drama, Biography, Romance & 2014 & 13 \\
 & {Water for Elephants}         & - & Drama, Romance & 2011 & 15 \\
 & {Wild Hogs}                   & - & Action, Adventure, Comedy & 2007 & 18 \\
\midrule
\multirow{34}{*}{KO} 
 & {200 Pounds Beauty}           & 미녀는 괴로워       & Comedy, Drama, Music & 2006 & 28 \\
 & {A Violent Prosecutor}        & 검사외전            & Action, Comedy, Crime & 2016 & 8 \\
 & {Battle for Incheon: Operation Chromite} & 인천상륙작전 & Action, Drama, History & 2016 & 9 \\
 & {Cold Eyes}                   & 감시자들            & Action, Crime, Thriller & 2013 & 13 \\
 & {Deranged}                    & 연가시              & Drama, Sci-Fi, Thriller & 2012 & 15 \\
 & {Exit}                        & 엑시트              & Comedy & 2019 & 19 \\
 & {Extreme Job}                 & 극한직업            & Comedy, Crime & 2019 & 1 \\
 & {Hide and Seek}               & 숨바꼭질            & Horror, Mystery, Thriller & 2013 & 3 \\
 & {Jeon Woochi}                 & 전우치              & Action, Adventure, Comedy & 2009 & 3 \\
 & {Marathon}                    & 말아톤              & Biography, Drama, Sport & 2005 & 20 \\
 & {May 18}                      & 화려한 휴가         & Drama, History & 2007 & 13 \\
 & {Miss Granny}                 & 수상한 그녀         & Comedy, Fantasy, Music & 2014 & 15 \\
 & {My Tutor Friend}             & 동갑내기 과외하기   & Action, Comedy, Romance & 2003 & 36 \\
 & {Northern Limit Line}         & 연평해전            & Drama, War & 2015 & 11 \\
 & {Ode to My Father}            & 국제시장            & Drama, War & 2014 & 11 \\
 & {Pandora}                     & 판도라              & Disaster, Action, Drama & 2016 & 29 \\
 & {Punch}                       & 완득이              & Comedy, Drama, Sport & 2011 & 25 \\
 & {Secret Reunion}              & 의형제              & Action, Drama, Thriller & 2010 & 24 \\
 & {Secretly, Greatly}           & 은밀하게 위대하게   & Drama, Action, Comedy & 2013 & 13 \\
 & {Silmido}                     & 실미도              & Action, Drama & 2003 & 24 \\
 & {Sunny}                       & 써니                & Comedy, Drama & 2011 & 31 \\
 & {Take Off}                    & 국가대표            & Comedy, Drama, Sport & 2009 & 31 \\
 & {The Attorney}                & 변호인              & Crime, Drama, History & 2013 & 7 \\
 & {The Berlin File}             & 베를린              & Spy, Action, Thriller & 2013 & 18 \\
 & {The Himalayas}               & 히말라야            & Adventure, Biography, Drama & 2015 & 5 \\
 & {The Neighbors}               & 이웃사람            & Thriller, Mystery & 2012 & 36 \\
 & {The Priests}                 & 검은 사제들         & Horror, Mystery, Thriller & 2015 & 9 \\
 & {The Roundup}                 & 범죄도시2           & Action, Crime, Thriller & 2022 & 21 \\
 & {The Thieves}                 & 도둑들              & Action, Comedy, Crime & 2012 & 31 \\
 & {Tidal Wave}                  & 해운대              & Action, Drama, Sci-Fi & 2009 & 17 \\
 & {Tunnel}                      & 터널                & Disaster, Drama & 2016 & 35 \\
 & {Veteran}                     & 베테랑              & Action, Comedy, Crime & 2015 & 6 \\
\bottomrule
\end{tabular}%
}
\caption{List of movies in \dataset. The genre (top three) and release year are sourced from IMDb. The dataset contains 60 movies (English 28 / Korean 32) spanning various genres.}
\label{tab:movies}
\end{table*}

\end{document}